\documentclass{article}

\usepackage[main, final]{neurips_2026}
\usepackage{graphicx}

\usepackage[utf8]{inputenc} 
\usepackage[T1]{fontenc}    
\usepackage{float}          
\usepackage{hyperref}       
\usepackage{url}            
\usepackage{booktabs}       
\usepackage{amsfonts}       
\usepackage{nicefrac}       
\usepackage{microtype}      
\usepackage[table]{xcolor}  
\DeclareRobustCommand{\rev}[1]{#1} 
\usepackage{amsmath}

\usepackage{xspace}
\usepackage{enumitem}
\usepackage{algorithm}
\usepackage{algpseudocode}
\usepackage{enumitem}
\usepackage{multirow}

\makeatletter
\newlength{\figcapskip}
\newlength{\tabcapskip}
\renewenvironment{table}
  {\setlength{\abovecaptionskip}{\z@}%
   \setlength{\belowcaptionskip}{\tabcapskip}%
   \@float{table}}
  {\end@float}
\makeatother

\title{InfCoiL: Coordinated Planner-Controller Learning for Closed-Loop Physics-Based Human-Object Interaction}

\author{%
  Yude Zou$^{1,2,*}$ \quad Yifei Yao$^{2,*}$ \quad Yiwei Hao$^{1,*}$ \quad Hanqing Wang$^{2}$ \quad Xing Gao$^{2,\dagger}$ \\
  $^{1}$Shanghai Jiao Tong University \quad $^{2}$Shanghai Artificial Intelligence Laboratory \\
}

\begin{document}

\maketitle
{\renewcommand{\thefootnote}{}\footnotetext{$^{*}$Equal contribution. \quad $^{\dagger}$Corresponding author: gaoxing@pjlab.org.cn.}}

\begin{abstract}
Text- and goal-conditioned human-object interaction (HOI) provides an intuitive interface for specifying high-level intents in physics-based character animation. Mapping such intents to physically plausible, contact-rich behaviors naturally calls for a generative planner and a physics-based controller, yet existing planner-controller systems either execute plans open-loop or close the loop while keeping the planner fixed, so execution feedback never reshapes what is planned. We present InfCoiL, a closed-loop framework for physics-based HOI from coordinated planner-controller learning. InfCoiL couples a morphology-aware flow-matching planner with a unified multi-morphology controller through a shared morphology interface in an autoregressive plan-and-execute loop, and alternately fine-tunes both components with reinforcement learning from the same physical rollouts. To optimize the flow-matching planner online, we introduce Semantic Flow Policy Optimization (SFPO), which stochasticizes selected denoising transitions with channel-wise semantic covariance over heterogeneous HOI features and assigns group-relative, sequence-level rewards from complete closed-loop executions. 
\rev{On held-out OMOMO sequences, InfCoiL outperforms open-loop hierarchical baselines in goal-reaching task success and physical tracking accuracy.} Project page: \rev{\url{https://infcoil.github.io/}}.
\end{abstract}

\section{Introduction}
\setlength{\parskip}{5.5pt plus 2pt}
Text- and goal-conditioned physics-based human-object interaction (HOI) is an important step toward controllable character animation and embodied agents. Language specifies high-level intents, while spatial goals ground object-centric interactions geometrically, together forming an intuitive interface for specifying diverse, physically executable interaction. 
Realizing this interface requires bridging high-level semantic intent to low-level physical control for executable contact-rich interactions.

Existing approaches address different parts of this hierarchy. Kinematic generative models, such as diffusion-based interaction generators~\cite{xu2023interdiff, li2024controllable, wang2026unleashing}, learn diverse motion manifolds from large-scale mocap data and support flexible, text-conditioned synthesis. Lacking physical constraints, though, they often suffer from physical implausibility such as human-object penetration and foot skating. Physics-based imitation policies~\cite{xu2025intermimic, xu2026interprior} improve dynamic plausibility by tracking reference motions in simulation, yet depend on prescribed references and cannot directly infer fine-grained physical behaviors from high-level semantic conditions. Hierarchical planner-controller systems combine these strengths by using a generative planner to produce references for a physical controller. However, existing hierarchies still lack closed-loop planner-controller co-adaptation: SimGenHOI~\cite{lin2025simgenhoi} aligns its generator via supervised fine-tuning on simulator-executed trajectories but still executes generated plans open-loop, whereas CLoSD~\cite{tevet2024closd} replans online from simulated states for humanoid motion but adapts only the tracking controller while keeping the diffusion planner fixed.

A natural next step is to close this loop, allowing the planner to recurrently condition on the physical states reached by the controller and revise future interaction plans accordingly. In HOI, however, such coordination is not simply achieved by feeding simulator states back to a generative planner. Mocap-trained planners generate morphology-dependent interaction references, whereas controllers are often specialized to specific simulated embodiments, making physical feedback difficult to interpret across morphologies. Meanwhile, planner outputs may fall outside the controller's executable distribution, and controller rollouts may visit physical states unseen during the planner's supervised training. These gaps are further amplified by contact-rich object dynamics, where small errors in contact timing, position, or object pose can quickly lead to cascading failures. Together, these issues reveal \textit{ a fundamental coordination problem between generative interaction planning and physical control.}

\begin{figure}[t]
    \centering
    \includegraphics[width=0.995\textwidth]{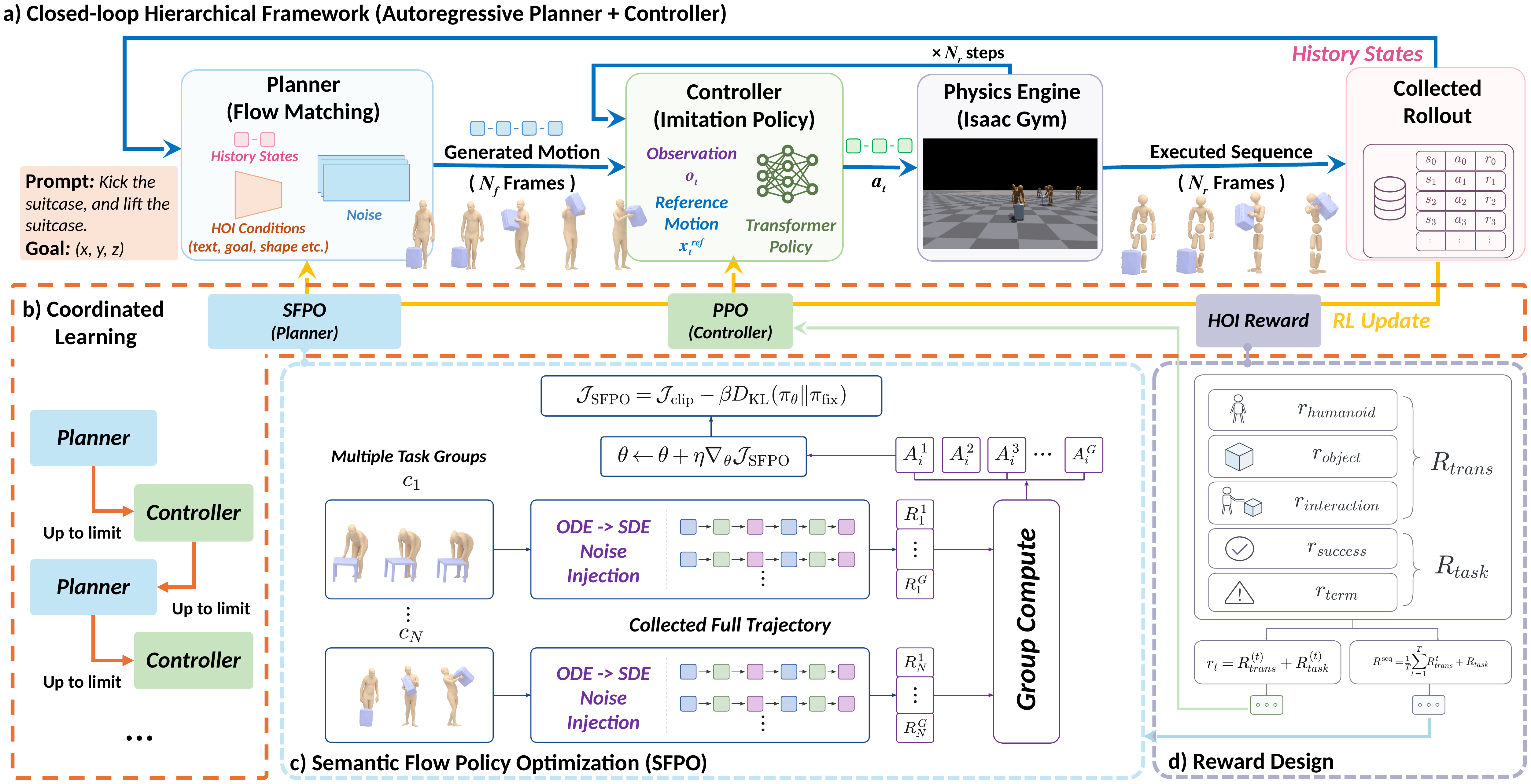}
    \caption{Overview of InfCoiL. (a) Closed-loop framework: the flow-matching planner generates an $N_f$-frame reference window, the controller executes its first $N_r$ frames in Isaac Gym, and the realized human, object, and contact states are fed back for receding-horizon replanning. (b) Coordinated learning: controller (PPO) and planner (SFPO) training alternate \rev{to narrow the gap between planner-generated references and what the controller can execute}. (c) SFPO: \rev{for each task condition, a group of plans is sampled with semantic noise injected during denoising and executed in closed loop. Their sequence rewards are normalized within the group into relative advantages that update the planner}. (d) Reward design: the controller is trained on stepwise transition and task rewards, and each plan is scored by the time-averaged sequence reward. The interaction term is used only in open-loop controller training.}
    \label{fig:maindiagram}
\end{figure}

To address this problem, we propose \textbf{InfCoiL}, a closed-loop framework for producing physics-based human-object \underline{in}teractions \underline{f}rom \underline{co}ord\underline{i}nated planner-controller \underline{l}earning, as shown in Fig.~\ref{fig:maindiagram}. InfCoiL closes three coupled loops between a generative planner and a physics-based controller: an interface loop that maps realized human-object-contact states back into the planner's conditioning across morphologies, an execution loop that replans from these states, and a learning loop that adapts both modules from the same physical rollouts. The first two loops are realized by \textit{an interface-compatible hierarchical architecture}. A morphology-aware flow-matching planner trained on diverse mocap data generates short-horizon interaction references from textual instructions and spatial goals, conditioned on the target morphology and the realized execution history. A unified multi-morphology controller, rather than separate embodiment-specific policies, executes these references across body shapes.

\rev{To narrow the gap between planner-generated references and what the controller can execute, the learning loop alternately fine-tunes the controller to track these references and the planner to generate references that lead to better execution. Online planner fine-tuning, however, requires transition likelihoods that deterministic flow sampling lacks, and exploration suited to HOI tokens, whose body, object, and contact channels differ in physical sensitivity and error tolerance. We therefore propose Semantic Flow Policy Optimization (SFPO), which turns a fixed subset of denoising steps into Gaussian policy steps with semantic, channel-wise covariance.} This yields tractable transition log-probabilities and channel-aware exploration while preserving efficient flow sampling. \rev{Together, SFPO and PPO controller fine-tuning align} the planner's generative distribution with the controller's physical execution capability. Our contributions are summarized as follows:\label{para:checklist1}
\begin{itemize}
    \item We formulate closed-loop physics-based HOI as a planner-controller coordination problem and propose \textbf{InfCoiL}, which closes interface, execution, and learning loops between a morphology-aware flow-matching planner and a unified multi-morphology controller.
    \item We propose SFPO, an online RL method for flow-matching HOI planners that stochasticizes selected denoising transitions with semantic channel-wise covariance and optimizes group-relative, sequence-level outcomes of complete closed-loop executions, and embed it in an alternating planner-controller learning scheme.
    \item \rev{Under a task-oriented evaluation on OMOMO~\cite{li2023object} sequences held out from training}, we show that InfCoiL improves task success over open-loop and SFT-aligned cascades, and provide component-wise ablations showing that replanning, controller learning, and planner learning address different failure sources, with planner learning most effective after the controller enlarges the executable region.
\end{itemize}

\setlength{\parskip}{5.5pt}
\section{Methodology}
\label{sec:method}

We consider text- and goal-conditioned physics-based human-object interaction (HOI), where a simulated humanoid interacts with an object according to semantic and spatial task specifications. At simulation step $t$, the physical state $s_t$ consists of humanoid, object, and contact states. The task context is defined as

\begin{equation}
\label{eq:task_context}
c_t=(\ell, \rev{\xi^p_t}, \rev{\xi^o}, \rev{\beta}, o, h_t),
\end{equation}
where $\ell$ is the textual instruction, \rev{$\xi^p_t$} denotes the local pelvis sub-goal for the current planning window, \rev{$\xi^o$} specifies the final object goal, \rev{$\beta$} represents the humanoid morphology using SMPL-X shape parameters~\cite{pavlakos2019expressive}, $o$ encodes object geometry with BPS features~\cite{prokudin2019efficient}, and $h_t$ summarizes recent physical execution history. Given $c_t$ and $s_t$, the system outputs low-level actions $a_t$ to realize the specified interaction while satisfying spatial goals and physical constraints.

InfCoiL decomposes physics-based HOI into a high-level generative planner $P_\theta$ and a low-level physical controller $\pi_\phi$, coupled through a closed-loop plan-and-execute interface, as detailed in Sec.~\ref{sec.framework}. The planner predicts an $N_f$-frame reference window $\hat{x}_{t:t+N_f-1}$, while the controller executes the first $N_r$ simulation steps before replanning, with $N_r \le N_f$. Throughout the method, $t$ denotes simulation time, $t'$ denotes an execution step within a replanning segment, $\tau \in [0,1]$ is reserved for flow-matching time, and $K$ denotes the number of flow denoising steps. To further align planning and control, InfCoiL jointly optimizes both components via coordinated reinforcement learning, mitigating the mismatch between generated references and executable behaviors (Sec.~\ref{sec.learning}).

\subsection{Closed-loop Planner-Controller Framework for HOI}
\label{sec.framework}

\subsubsection{Closed-loop versus Open-loop Design}

InfCoiL closes the planner-controller loop at the reference-trajectory level. 
In a conventional open-loop cascade, a generative model produces a complete kinematic interaction sequence, which is then tracked by a physics-based controller throughout the episode. 
This assumes that the simulated humanoid, object, and contacts remain close to the planned trajectory. 
However, contact-rich HOI easily violates this assumption: small errors in contact timing, object pose, root progress, or tracking accuracy accumulate over time and make the remaining reference physically infeasible.

InfCoiL therefore adopts a receding-horizon replanning strategy, where the planner iteratively generates short-horizon reference windows and refines future plans using the controller's realized physical states. Every $N_r$ simulation steps, the planner $P_\theta$ predicts
$\hat{x}_{t:t+N_f-1}=P_\theta(c_t)$,
where $c_t$ is the task context defined in Eq.~\eqref{eq:task_context}. The controller $\pi_\phi$ executes the next segment by tracking these references,
$a_{t'}=\pi_\phi(s_{t'},\hat{x}_{t'})$ for $t'\in[t,t+N_r)$, after which the resulting body-object-contact states are fed back to update $c_{t+N_r}$. This reference-level closed loop preserves the diversity and goal controllability of kinematic generation while enabling future plans to adapt to physical execution. Additional implementation details are provided in Appendix~\ref{app:implementation_details}.

\subsubsection{Morphology-aware Autoregressive Planner}

The high-level planner synthesizes short-horizon HOI references using a morphology-aware flow-matching model~\cite{lipman2022flow} trained on multi-subject, multi-object interaction data~\cite{li2023object}.

Given task conditions and recent execution history, it recurrently predicts a short-horizon kinematic reference window for the controller to track.
Each frame is represented as a structured HOI token containing body joints, hand joints, global body rotations, object translation and rotation, and hand contact labels. 
The planner is conditioned on text embeddings, local pelvis and object goals, object BPS features, temporal progress variables, fixed prefix frames, and a 16-dimensional SMPL-X shape code. 
The morphology condition allows the planner to adapt interaction references to different body shapes, since the same object-centric task may require different reaching poses, contact locations, and root motions across humanoid embodiments.

During training, we interpolate between a clean reference window $x_0$ and Gaussian noise $\epsilon$,
\begin{equation}
    x_\tau = \tau \epsilon + (1-\tau)x_0,\qquad \tau \in (0,1],
\end{equation}
and train the planner to predict the clean state $\hat{x}_0 = f_\theta(x_\tau,\tau,c),$
where $c$ denotes the full conditioning context, including text, goals, object geometry, morphology, progress, and prefix states. 
The supervised planning objective is
\begin{equation}
    \mathcal{L}_{\mathrm{plan}}
    =
    \|f_\theta(x_\tau,\tau,c)-x_0\|_2^2
    + \lambda_{\mathrm{obj}}\mathcal{L}_{\mathrm{obj}}
    + \lambda_{\mathrm{fk}}\mathcal{L}_{\mathrm{fk}},
\end{equation}
where $\mathcal{L}_{\mathrm{obj}}$ supervises object keypoints and $\mathcal{L}_{\mathrm{fk}}$ supervises forward-kinematic hand and foot positions, and $\lambda_{*}$ denotes corresponding hyperparameter. 
These auxiliary terms encourage the generated references to preserve object motion consistency and accurate end-effector geometry, which are critical for contact-rich HOI.

At inference time, the planner samples a reference window by integrating the induced velocity
\begin{equation}
    v_\theta(x_\tau,\tau,c)=\frac{x_\tau-f_\theta(x_\tau,\tau,c)}{\tau}
    \label{eq:velocity}
\end{equation}
from noise to data, while clamping the prefix frames from recent physical execution. 
Since $x_\tau$ is linear in $x_0$ and $\epsilon$, Eq.~\eqref{eq:velocity} is the velocity induced by the clean-state prediction, so the trained network is exactly the one being integrated. Predicting $x_0$ rather than $v=\epsilon-x_0$ regresses the structured HOI state instead of a noise-mixed target and avoids the $1/\tau^2$ reweighting of a velocity-space loss~\cite{li2025back}. The clean-state target directly supervises the positions, rotations, object states, and contact labels tracked by the controller, and Appendix~\ref{app:xv_ablation} shows that it reduces foot sliding and velocity error of the generated references compared with velocity prediction.

\subsubsection{Multi-morphology Controller}
The low-level controller provides a unified executable interface for morphology-conditioned planner references. 
This interface is essential for closed-loop HOI: since the planner generates references for the target humanoid morphology, the controller must execute them directly rather than retarget all motions to a canonical body. 
Unlike prior imitation frameworks such as InterMimic~\cite{xu2025intermimic}, which consolidate embodiment-specific policies into a shared student with a unified output representation, InfCoiL learns a morphology-conditioned controller that adapts its tracking behavior to different SMPL body shapes while remaining compatible with planner-generated references.

We initialize from morphology-specific InterMimic-style teachers trained for individual subjects and distill them into a single student controller through the distillation-to-RL curriculum of InterMimic~\cite{xu2025intermimic}. 
Each environment is assigned a subject-object pair, and the corresponding teacher labels every visited state with its action mean. Rollouts are initially driven by the teacher and gradually handed over to the student by mixing teacher and student actions with a decaying probability, so that expert labels are increasingly collected on student-induced states. After critic warm-up, PPO actor updates are enabled when a critic-reliability criterion is satisfied, with the actor weight gradually increased and the imitation weight reduced, allowing the student to inherit subject-specific tracking skills while adapting to its own induced state distribution.

The student controller is conditioned on the same 16-dimensional SMPL-X shape code $\beta$ used by the planner, together with a reference-tracking context \rev{$\hat{x}$} over four lookahead steps.
These inputs encode both body morphology and near-future interaction targets, enabling morphology-adaptive tracking under contact-rich dynamics. 
A compact transformer policy embeds the inputs and outputs a Gaussian action mean in the 153-dimensional action space. 
This yields a shared yet morphology-aware controller that supports closed-loop execution of planner-generated references across diverse humanoid shapes and object interactions. 

Extending reference state initialization~\cite{peng2018deepmimic} and diverse resets~\cite{yin2026emergent}, we further introduce contact-aware state initialization (CSI) to focus controller pretraining on critical HOI phases. 
For each demonstration, reset weights are computed from hand-contact label transitions, so that rollouts are more frequently initialized near contact onset and release. 
This exposes the controller to diverse and challenging body-object configurations where tracking errors are most consequential, reducing the need to discover rare contact states through long rollouts and improving contact handling under planner-generated references. Appendix~\ref{app:intermimic_tracking_eval} shows that removing CSI lowers SR by up to 25 points.

\subsection{Coordinated Planner-Controller Learning}
\label{sec.learning}
The closed-loop framework in Sec.~\ref{sec.framework} enables physical execution feedback for future planning, but does not by itself resolve the planner-controller distribution mismatch, \emph{i.e.}, generated references may be hard to execute and controller rollouts may reach states outside the planner's pretraining distribution. To bridge this shift, InfCoiL introduces coordinated planner-controller learning with shared closed-loop rollout feedback. The controller is optimized to execute planner-generated references under simulation dynamics, while the planner is fine-tuned to improve downstream execution outcomes, forming a coupled optimization between generative planning and physical control. We realize this coordination with SFPO for the planner, PPO-based controller optimization, and interaction-centric rewards. Together, these updates align the planner's generative distribution with the controller's executable behavior distribution.

\subsubsection{SFPO for Planner Learning}
\label{sec:planner_learning}

We fine-tune the flow-matching planner with physical execution feedback from the closed-loop controller. 
This is nontrivial because rectified-flow sampling is typically deterministic, and thus does not directly provide the transition likelihoods required by policy-gradient objectives~\cite{zhang2025reinflow, liu2025flow, mcallister2025flow}. 
Moreover, HOI references are high-dimensional and heterogeneous, containing body motion, object states, and contact labels, making naive stochastic perturbations unstable and inefficient.

To address this, SFPO builds on the stochastic denoising path of Flow-GRPO~\cite{liu2025flow} rather than the flow-matching likelihood surrogate of FPO~\cite{mcallister2025flow}. Let $z_k=x_{\tau_k}$ ($k=1,\dots,K+1$) be the states of the $K$-step Euler sampler, from $\tau_1=1$ to $\tau_{K+1}=0$ with $\Delta\tau_k=\tau_{k+1}-\tau_k<0$. \rev{SFPO turns a fixed subset $\mathcal{S}$ of $S\le K$ steps into Gaussian policy steps by Euler--Maruyama discretization of the marginal-preserving flow SDE and keeps the deterministic Euler update at all other steps}:
\begin{equation}
\pi_\theta(z_{k+1}\mid z_k,c_t)=\mathcal{N}\big(\mu_{\theta,k},\,(-\Delta\tau_k)\Gamma_{\tau_k}\Gamma_{\tau_k}^{\top}\big),\qquad
\mu_{\theta,k}=z_k+\bar v_\theta(z_k,\tau_k,c_t)\,\Delta\tau_k,\qquad k\in\mathcal{S},
\label{eq:sfpo_policy}
\end{equation}
where $\bar v_\theta$ is a score-free drift computed from the velocity in Eq.~\eqref{eq:velocity}. We use $K=10$ and make only the final step stochastic ($\mathcal{S}=\{K\}$), which \rev{reduces log-probability recomputation to a single denoising step per plan}. To match exploration to heterogeneous HOI tokens, the diffusion coefficient is semantic rather than isotropic, $\Gamma_\tau=\lambda_{\mathrm{n}}\sqrt{\tau/(1-\tau)}\operatorname{Diag}(\rev{\kappa})$: fixed hand-set scales $\rev{\kappa}\in\mathbb{R}_+^D$ perturb object translation most and rotations least, and $\lambda_{\mathrm{n}}$ sets their global level.

Planner updates are on-policy. A frozen snapshot $\theta_{\mathrm{old}}$ generates the plans and stores their transition log-probabilities, which are recomputed under the current $\theta$ on the same denoising path. For a task condition $c_i$, we execute $G$ closed-loop rollouts, and every replanning event $t$ of the $g$-th rollout forms one training sample with ratio $\rho_\theta^{i,g,t}=\exp(\bar\ell_\theta^{\,i,g,t}-\bar\ell_{\theta_{\mathrm{old}}}^{\,i,g,t})$, where $\bar\ell$ denotes the transition log-probability averaged over the generated elements and clipped for numerical stability. \rev{Since each plan is scored only by the executed trajectory it leads to}, all samples of a rollout share one group-relative advantage~\cite{shao2024deepseekmath} computed from the \rev{sequence reward $R_i^g$ of rollout $g$ (Eq.~\eqref{eq:sequence_reward})}:
\begin{equation}
A_i^g=\operatorname{clip}\Big(\frac{R_i^g-\rev{\bar R_i}}{\rev{\varsigma_i}},\,-A_{\max},\,A_{\max}\Big),
\end{equation}
where \rev{$\bar R_i$ and $\varsigma_i$ are the mean and standard deviation of $R_i^g$ over the valid rollouts of group $i$, and} groups with near-zero reward variance are skipped. The planner then maximizes the clipped surrogate \rev{$\mathcal{J}_{\mathrm{clip}}$} with an optional fixed-reference penalty
\begin{equation}
\begin{aligned}
\rev{\mathcal{J}_{\mathrm{clip}}}&=\mathbb{E}_{i,g,t}\Big[\min\big(\rho_\theta^{i,g,t}A_i^g,\ \operatorname{clip}(\rho_\theta^{i,g,t},1-\epsilon_{\mathrm{c}},1+\epsilon_{\mathrm{c}})A_i^g\big)\Big],\\
\mathcal{J}_{\mathrm{SFPO}}&=\rev{\mathcal{J}_{\mathrm{clip}}-\lambda_{\mathrm{KL}}\,\mathbb{E}_{i,g,t,\,k\in\mathcal{S}}\Big[D_{\mathrm{KL}}\big(\pi_\theta(\cdot\mid z_k,c_t)\,\Vert\,\pi_{\mathrm{fix}}(\cdot\mid z_k,c_t)\big)\Big]}.
\end{aligned}
\label{eq:sfpo_objective}
\end{equation}
Here $\pi_{\mathrm{fix}}$ is a frozen pretrained planner, and \rev{$\lambda_{\mathrm{KL}}$ weights this optional penalty}. We set \rev{$\lambda_{\mathrm{KL}}=0$} in the reported experiments, \rev{so that $\mathcal{J}_{\mathrm{SFPO}}=\mathcal{J}_{\mathrm{clip}}$}.
Appendix~\ref{sec:appendix_planner_sde} gives the derivation of $\bar v_\theta$, the per-channel values of \rev{$\kappa$}, the exact log-probability aggregation, and all hyperparameters.

\subsubsection{Human-Object Interaction-Centric Reward}
\label{sec:hoi_reward}
We design an HOI-centric reward to guide both local reference tracking and episode-level task completion, following the interaction-aware reward principles of InterMimic~\cite{xu2025intermimic}. 
The reward consists of a stepwise transition reward $R_{\mathrm{trans}}^t$ and a task reward $R_{\mathrm{task}}$ triggered only at the end of the episode.

The transition reward measures the consistency between the current simulated state and the reference interaction at each simulation step:
\begin{equation}
R_{\mathrm{trans}}^t = \rev{w_{\mathrm{h}}} r_{\mathrm{humanoid}}^t + \rev{w_{\mathrm{o}}} r_{\mathrm{object}}^t + \rev{w_{\mathrm{int}}} r_{\mathrm{interaction}}^t.
\label{eq:transition_reward}
\end{equation}
Here, $r_{\mathrm{humanoid}}^t$ and $r_{\mathrm{object}}^t$ are exponential tracking rewards based on the mean per-joint position error for the humanoid and the object center position error, respectively. The term $r_{\mathrm{interaction}}^t$ measures the interaction consistency between the human and the object based on hand-object contact labels, contact point distances, and interaction graph errors.

The interaction term $r_{\mathrm{interaction}}^t$ is used only in the open-loop reinforcement learning of the controller and is disabled during closed-loop training. When the planner and the controller adapt to each other, rewarding interaction-graph consistency invites reward hacking toward trajectories with weak or negligible contact, which lowers task success. Closed-loop training therefore relies on the humanoid and object tracking rewards and the terminal task reward.

The task reward is triggered only at the end of the episode:
\begin{equation}
R_{\mathrm{task}} = \rev{w_{\mathrm{succ}}} R_{\mathrm{success}} + w_{\mathrm{term}} R_{\mathrm{term}}.
\label{eq:task_reward}
\end{equation}
In this formulation, $R_{\mathrm{success}}$ rewards episodes that successfully complete the target human-object interaction. The term $R_{\mathrm{term}}$ \rev{penalizes episodes that terminate early due to infeasible humanoid motions or excessive object deviation from the reference trajectory, or that reach the timeout without completing the task}. The overall reward therefore combines dense local tracking signals with sparse task-level evaluation, allowing the controller and planner to improve physical execution while remaining aligned with the intended HOI objective.
 
\subsubsection{Execution Feedback and Group-Relative Advantage}
\label{sec:group_advantage}

The controller and the planner share these reward components but use them differently. The controller is a stepwise policy trained with PPO and a learned value function \rev{to maximize the expected return of the per-step reward $R_{\mathrm{trans}}^t$, with $R_{\mathrm{task}}$ added at the final step}. The planner makes low-frequency decisions over long horizons whose quality is revealed only by the complete execution. \rev{Each rollout of length $T$ is therefore scored by a sequence reward that averages the transition reward over time, so that the planner cannot gain reward merely by prolonging rollouts}:
\begin{equation}
R^{\mathrm{seq}} = \frac{1}{T} \sum_{t=1}^{T} R_{\mathrm{trans}}^t + R_{\mathrm{task}}.
\label{eq:sequence_reward}
\end{equation}

As described in Sec.~\ref{sec:planner_learning}, planner optimization in SFPO is critic-free and uses group-relative advantages. Rollouts in a group share the interaction conditions, such as the start state and task goals, and differ only in the semantic denoising noise. Normalizing the sequence scores $R_i^g$ within each group therefore compares plans under identical conditions and removes reward offsets caused by task difficulty or object scale, which encourages references that the controller can execute. \rev{The reward terms,} group construction, valid-set masking, and terminal advantage assignment are detailed in Appendix~\ref{sec:appendix_group_advantage}.

\subsubsection{Coordinated Learning Paradigm}
\label{sec:coordinated_learning}

Optimizing either component in isolation is insufficient. A controller trained alone adapts to fixed planner references but cannot change the motion generation itself, which bounds task success. A planner updated against a frozen controller instead explores references that the controller cannot execute, so it receives little useful feedback. InfCoiL therefore alternates planner and controller updates. After supervised pretraining of the planner and open-loop distillation \rev{and RL} of the controller, the controller is first fine-tuned with PPO on closed-loop rollouts. Once its tracking quality or success rate plateaus, the controller is frozen and the planner is updated with SFPO until its sequence-level scores plateau, after which training returns to the controller. The controller phase thus resolves the physical executability of current references, and the planner phase moves references toward what the controller can stably track. Algorithm~\ref{alg:coordinated} and Appendix~\ref{sec:appendix_coordinated_learning} give the full procedure.

\setlength{\parskip}{5.5pt plus 2pt}
\section{Experiments}
\label{sec:Exp}

\subsection{Evaluation Settings}
\label{metrics}
\textbf{Metrics and protocol.}
We evaluate all methods on a morphology-balanced re-split of OMOMO~\cite{li2023object}, which splits the sequences of every subject with the same train/test ratio. Planner training, controller training, and coordinated fine-tuning use only the training sequences, and all results are reported on the held-out test sequences. All methods share the same Isaac Gym environment, humanoid/object assets, physical parameters, and deterministic controller actions. Each test sequence is evaluated three times and metrics are averaged. SR@$d$ counts an episode as successful when both the pelvis and the object end within $d$ m of their goals. SR denotes SR@0.3 unless stated otherwise, and Table~\ref{tab:planner_rl} and the appendix additionally report SR@0.2, SR@0.1, and their mean (Avg.\ SR). We also report pelvis/object goal errors and humanoid/object tracking errors $E_h$/$E_o$, all in centimeters. This protocol requires reaching the task goals, unlike the reference-tracking protocol of InterMimic~\cite{xu2025intermimic}, which tracks references on a teacher-supported subset, counts episodes within a 0.5 m deviation as successful, and reports the best sampled rollout. Appendix~\ref{app:intermimic_gap} explains how this difference affects the InterMimic student results, and Appendix~\ref{app:experimental_protocol} gives further protocol and metric details.

\begin{figure}[t]
  \centering
  \includegraphics[width=\linewidth,trim=0 3.5 0 0,clip]{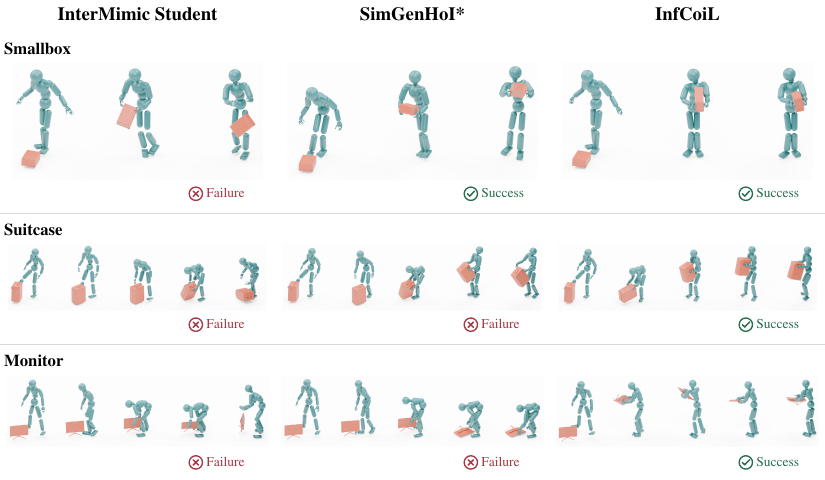}
    \caption{Qualitative comparison on three OMOMO sequences. The instructions, from top to bottom, are `kick and lift smallbox', `kick and lift suitcase', and `lift and rotate monitor'. Green checks and red crosses denote recorded rollout success and failure. In the small-box example, InterMimic student lifts the box but loses its grasp; this rollout is a failure.}
  \label{fig:expcomp}
\end{figure}

\textbf{Baselines and variants.}
We compare InfCoiL with two open-loop baselines and four internal variants. In all cases, the controller tracks reference trajectories produced by a high-level planner and the key difference is whether these references are generated once in an open-loop manner or repeatedly generated in a closed-loop manner with execution feedback. InterMimic~\cite{xu2025intermimic} student evaluates the distilled multi-morphology controller under the one-shot planner-reference protocol. SimGenHOI*~\cite{lin2025simgenhoi} follows an open-loop cascade with two-stage supervised fine-tuning (SFT) with the same network architecture of planner and controller as ours for fair comparison: the planner is first fine-tuned on trajectories produced by controller tracking, and the controller is then fine-tuned on trajectories generated by the updated planner. InfCoiL open-loop uses our planner and controller with one-shot reference generation and no closed-loop replanning. The closed-loop variant w/o FT P \& C instead uses closed-loop online generation with receding-horizon feedback, but freezes both modules. Finally, w/o FT C fine-tunes only the planner, w/o FT P fine-tunes only the controller, and full InfCoiL alternates controller (C) and planner (P) fine-tuning phases. We report InfCoiL (C$\to$P), with one controller phase followed by one planner phase, and InfCoiL (C$\to$P$\to$C), which adds a further controller phase. Unless otherwise stated, InfCoiL refers to the latter.

\begin{table}[t]
\caption{Task-oriented evaluation on the re-split OMOMO test set. Each sequence is evaluated three times and metrics are averaged. C and P denote controller (PPO) and planner (SFPO) fine-tuning phases: InfCoiL (C$\to$P) uses 10k controller iterations followed by 500 planner iterations, and InfCoiL (C$\to$P$\to$C) adds 4k further controller iterations.}
\label{tab:hoi_main_results}
\centering 

\small
\setlength{\tabcolsep}{10pt}
\begin{tabular}{@{}lccccc@{}} 
\toprule  
Method & SR@0.3$\uparrow$ & $E_h\downarrow$ & $E_o\downarrow$ & Pelvis Goal$\downarrow$ & Object Goal$\downarrow$ \\
\midrule
InterMimic student (open-loop) & 24.62 & 9.23 & 16.89 & 90.39 & 111.89 \\
SimGenHOI* (open-loop + SFT) & 52.14 & 9.69 & 18.00 & 50.12 & 68.15 \\
\midrule
InfCoiL open-loop 
& 49.23 & 9.40 & 18.08 & 57.15 & 75.51 \\

InfCoiL w/o FT P \& C 
& 56.09 & \textbf{7.25} & 16.10 & 46.89 & 69.83 \\

InfCoiL w/o FT C 
& 56.98 & 7.75 & 16.36 & 47.02 & 70.60 \\

InfCoiL w/o FT P
& 73.37 & 8.44 & 14.67 & 35.72 & 49.45  \\
\midrule
InfCoiL (C$\to$P)
& 77.24 & 8.90 & 14.76 & 32.90 & 45.37  \\

InfCoiL (C$\to$P$\to$C)
& \textbf{81.19} & 8.87 & \textbf{14.21} & \textbf{31.31} & \textbf{41.51}  \\
\bottomrule
\end{tabular}

\end{table}

\subsection{Results and Analysis}
\label{sec:results}

Table~\ref{tab:hoi_main_results} shows that InfCoiL substantially improves task completion over open-loop baselines. Compared with InterMimic student, InfCoiL raises SR from 24.62\% to 81.19\% and reduces pelvis/object goal errors from 90.39/111.89 cm to 31.31/41.51 cm. Compared with SimGenHOI* (open-loop + SFT), InfCoiL increases SR from 52.14\% to 81.19\%, while reducing pelvis/object goal errors from 50.12/68.15 cm to 31.31/41.51 cm. These gains show that closed-loop planner-controller learning improves task completion beyond one-shot planner-reference generation. InfCoiL also improves physical tracking quality over  open-loop baselines while achieving higher task success. Compared with InterMimic student and SimGenHOI*, InfCoiL reduces $E_h$ from 9.23/9.69 cm to 8.87 cm and $E_o$ from 16.89/18.00 cm to 14.21 cm. This indicates that closed-loop  learning improves not only goal reaching, but also the executability of planner-generated references. As shown in Fig.~\ref{fig:expcomp}, InfCoiL achieves more stable contacts and physically plausible object interactions\rev{, and Appendix~\ref{app:failure_analysis} analyzes the failure modes}. Furthermore, Figs.~\ref{fig:random_sampled_beta_obj} and~\ref{fig:random_sampled_beta_obj_part2} demonstrate the generalization to novel object-humanoid morphology combinations.

\subsection{Framework Ablation}
\label{sec:ablation}

\textbf{Component summary.} Each component of InfCoiL is isolated by a controlled comparison. Closed-loop replanning alone raises SR from 49.23\% to 56.09\%, and lowers both goal errors and tracking errors, showing that replanning corrects part of the mismatch between generated references and realized physical states. Controller adaptation (w/o FT P) adds a further 17.28 points, and alternating updates are compared with controller-only training at a matched controller budget below. SFPO is compared with alternative planner RL objectives in Sec.~\ref{sec:planner_rl_ablation}, online RL is compared with supervised fine-tuning of the planner in Appendix~\ref{app:sft_vs_rl}, and removing contact-aware state initialization lowers SR under the reference-tracking protocol of Appendix Table~\ref{tab:teacher_student_results}.

\textbf{Effect of coordinated learning.} Table~\ref{tab:hoi_main_results} shows that fine-tuning only one component exposes the asymmetry between planner and controller adaptation. Planner-only fine-tuning (w/o FT C) provides only a small SR gain over the frozen closed-loop system, from 56.09\% to 56.98\%, suggesting that a frozen controller restricts the executable support and thus the useful reward signal for planner learning. Controller-only fine-tuning (w/o FT P) is much stronger, raising SR to 73.37\% and reducing object-goal error to 49.45~cm, but it still leaves the planner distribution fixed. Alternating the two phases yields consistent gains: a subsequent SFPO planner phase raises SR to 77.24\% (C$\to$P), and a further controller phase reaches 81.19\% (C$\to$P$\to$C) while reducing pelvis/object goal errors to 31.31/41.51 cm. In contrast, the matched-budget comparison in Appendix~\ref{app:matched_budget} shows that continuing controller-only training to the same 14k controller iterations lowers SR to 71.99\%, indicating that the gain comes from planner-controller co-adaptation rather than additional controller updates. Although the lowest humanoid tracking error $E_h$ is attained by the frozen closed-loop variant, full InfCoiL achieves the best task success, goal accuracy, and object tracking error, suggesting that coordinated learning optimizes executable task completion rather than tracking fidelity alone.

\subsection{Planner RL Ablation}
\label{sec:planner_rl_ablation}

\begin{table}[t]
\caption{Planner RL ablation. All methods start from the same controller-adapted checkpoint (InfCoiL w/o FT P in Table~\ref{tab:hoi_main_results}) and update the planner with the controller frozen. SFPO with $\lambda_{\mathrm{n}}=0.1$ corresponds to InfCoiL (C$\to$P). BC is behavior-cloning anchor and Avg.\ SR averages the three thresholds, where errors are in cm.}
\label{tab:planner_rl}
\centering
\small
\setlength{\tabcolsep}{3.5pt}
\begin{tabular}{@{}lcccccccc@{}}
\toprule
Method & SR@0.3$\uparrow$ & SR@0.2$\uparrow$ & SR@0.1$\uparrow$ & Avg.\ SR$\uparrow$ & $E_h\downarrow$ & $E_o\downarrow$ & Pelvis$\downarrow$ & Object$\downarrow$ \\
\midrule
Start point (w/o FT P) & 73.37 & 67.96 & 47.54 & 62.96 & 8.44 & 14.67 & 35.72 & 49.45 \\
\midrule
FPO++~\cite{yi2026flow} & 12.83 & 9.69 & 4.20 & 8.91 & 14.29 & 22.25 & 120.03 & 136.34 \\
FPO++ + BC & 73.45 & 67.64 & 46.57 & 62.55 & 8.52 & \textbf{14.53} & 37.44 & 51.23 \\
ReinFlow PPO~\cite{zhang2025reinflow} & 70.06 & 50.20 & 14.21 & 44.82 & 12.18 & 19.07 & 44.52 & 62.82 \\
ReinFlow PPO + BC & 73.61 & 67.55 & 44.31 & 61.82 & \textbf{8.42} & 14.86 & 36.78 & 49.40 \\
Flow-GRPO~\cite{liu2025flow} (isotropic) & 22.52 & 17.59 & 6.86 & 15.66 & 19.02 & 22.79 & 95.75 & 120.75 \\
\midrule
SFPO ($\lambda_{\mathrm{n}}=0.05$) & 74.66 & 68.77 & 47.22 & 63.55 & 8.77 & \textbf{14.53} & 34.59 & 47.33 \\
SFPO ($\lambda_{\mathrm{n}}=0.2$) & 75.14 & 69.09 & 44.15 & 62.79 & 8.98 & 16.42 & 36.89 & 48.48 \\
SFPO ($\lambda_{\mathrm{n}}=0.1$) & \textbf{77.24} & \textbf{72.07} & \textbf{48.59} & \textbf{65.97} & 8.90 & 14.76 & \textbf{32.90} & \textbf{45.37} \\
\bottomrule
\end{tabular}
\end{table}

We compare SFPO with planner-side reinforcement learning baselines under the same closed-loop setting. All methods start from the same controller-adapted checkpoint, keep the controller frozen, and update the planner with the same number of iterations. The baselines cover critic-based single-step updates, namely FPO++~\cite{yi2026flow}, an improved variant of FPO~\cite{mcallister2025flow}, and ReinFlow PPO~\cite{zhang2025reinflow}, each with and without a behavior-cloning (BC) anchor, as well as Flow-GRPO~\cite{liu2025flow} with isotropic noise. Implementation details are provided in Appendix~\ref{app:planner_rl_ablation}.

Table~\ref{tab:planner_rl} shows that critic-based updates do not improve on this strong starting point: without a BC anchor, FPO++ and ReinFlow PPO degrade to 12.83\% and 70.06\% SR@0.3, and with the anchor they at most preserve the initial policy (73.45\% and 73.61\% vs.\ 73.37\%). Making the flow sampler stochastic with isotropic noise is also insufficient, as Flow-GRPO drops to 22.52\%. In contrast, SFPO improves SR@0.3 over the starting point at every tested noise level ($\lambda_{\mathrm{n}}\in\{0.05, 0.1, 0.2\}$, 74.66--77.24\%), and the default $\lambda_{\mathrm{n}}=0.1$ achieves the best success at all thresholds together with the lowest goal errors, while a larger noise level trades strict-threshold success (SR@0.1) for coarse-threshold gains. These results indicate that channel-wise exploration matched to heterogeneous HOI tokens is what makes sequence-level planner RL stable in this setting. Appendix~\ref{app:sft_vs_rl} further compares online RL with supervised fine-tuning of the planner.

\section{Related Works}

\textbf{Generative Models for Human-Object Interaction.}
Kinematic generative models have become a dominant approach for synthesizing human-object interactions (HOI). 
Diffusion-based methods incorporate various conditions to support object-guided manipulation and text-controllable interaction synthesis~\cite{xu2023interdiff,li2023object,diller2024cg,li2024controllable}. 
Subsequent work improves conditioning and representation design through semantic-geometric context~\cite{cong2025semgeomo}, object relations~\cite{xue2025guiding}, kinematic-chain structures~\cite{zeng2025chainhoi}, unseen-object generalization~\cite{li2025genhoi}, and interaction dynamics~\cite{wu2025hoi}. 
Other studies accelerate or refine the generative process itself, including asynchronous denoising for classifier-free interaction guidance~\cite{wang2026unleashing} and consistency-based motion generators for fast inference~\cite{dai2024motionlcm,jiang2025motionpcm,zou2026infbagel}. 
However, flow-based HOI generation remains underexplored. Moreover, existing generators are typically trained without physical execution feedback, limiting their ability to adapt plans to the states  reached by the controller.

\textbf{Physics-based Human-Object Interaction.}
Physics-based HOI methods aim to execute interactions under contact-rich dynamics rather than only synthesize kinematic trajectories. A direct paradigm is to learn task policies with reinforcement learning from manually designed rewards, which can produce executable behaviors for sports, games, and object manipulation but often depends on task-specific engineering and lacks demonstration-level realism~\cite{merel2020catch, liu2018learning, xu2024humanvla}. Imitation-based methods reduce this burden by tracking human-object demonstrations in simulation, enabling contact-rich and reusable interaction skills~\cite{wang2023physhoi, wang2025skillmimic, yu2025skillmimic}. However, these skills remain bounded by the objects, motions, and interaction modes covered by demonstrations, limiting their ability to synthesize diverse long-horizon behaviors from fine-grained textual and spatial instructions. Recent large-scale methods improve coverage by learning whole-body controllers from broader and imperfect mocap interaction data, including unified controllers for heterogeneous objects and sparse-intention-conditioned control~\cite{xu2025intermimic, xu2026interprior}. These methods ground HOI in simulation, but they remain less expressive than text-conditioned generative planners in semantic controllability and motion diversity. Hierarchical systems further combine interaction generators with imitation policies to connect planning and physical tracking~\cite{lin2025simgenhoi}, but they still generate static references before tracking, leaving the planner essentially open-loop. In contrast, InfCoiL addresses this gap with a closed-loop planner-controller framework, where the planner repeatedly conditions on language, goals, morphology, object states, and recent physical feedback, and coordinated reinforcement learning adapts both modules to each other.

\section{Conclusion and Limitation}
\label{sec:conclusion}
This work studies how high-level generative interaction planning can be coordinated with low-level physical control for text- and goal-conditioned human-object interaction. A series of validations on the OMOMO benchmark shows that the closed-loop planner-controller framework enables adaptive replanning from physical execution feedback, while coordinated learning further aligns motion-capture-based generation with executable simulated behavior. \rev{The ablations also indicate when each component matters: closed-loop replanning corrects recoverable drift from the generated references, controller adaptation enlarges the physically executable region, and planner learning is most effective once this region has expanded, where it raises the final task-success ceiling.} Together, these findings suggest that physics-based HOI animation requires not only diverse and controllable generative priors, but also their closed-loop adaptation to the physical controller and contact-rich execution dynamics.

\rev{While InfCoiL demonstrates strong performance in simulation, several limitations remain. First, InfCoiL is a physics-based character animation system. All experiments use simulated SMPL-X humanoids, whose shape variation is not equivalent to the kinematics, actuation, and sensing of real humanoid robots, so real-robot deployment would require adaptation to hardware constraints, sensing noise, and real contact dynamics. Second, the current setting assumes structured object states and goal specifications, which offers a controlled testbed for studying planner-controller coordination but leaves perception-driven inputs and more complex scenes to future work. Third, the effective semantic range of language conditioning follows the OMOMO/CHOIS annotations, and we have not evaluated paraphrased or out-of-distribution instructions. Appendix~\ref{app:future_directions} discusses possible extensions.}

\setlength{\parskip}{5.5pt}
\begin{ack}
This work was supported in part by the National Natural Science Foundation of China under Grant 62401367 and in part by Shanghai Artificial Intelligence Laboratory.
\end{ack}

\small
\bibliography{main}
\bibliographystyle{unsrt}
\clearpage

\section*{NeurIPS Paper Checklist}

\begin{enumerate}

\item {\bf Claims}
    \item[] Question: Do the main claims made in the abstract and introduction accurately reflect the paper's contributions and scope?
    \item[] Answer: \answerYes{}
    \item[] Justification: The paper’s contributions and scope are specifically claimed in the abstract and introduction. Our key contributions are summarized in the last paragraph of \rev{Section~\ref{para:checklist1}}.

\item {\bf Limitations}
    \item[] Question: Does the paper discuss the limitations of the work performed by the authors?
    \item[] Answer: \answerYes{}
    \item[] Justification: \rev{We discuss the limitations of our work in Section~\ref{sec:conclusion}.}

\item {\bf Theory assumptions and proofs}
    \item[] Question: For each theoretical result, does the paper provide the full set of assumptions and a complete (and correct) proof?
    \item[] Answer: \rev{\answerNA{}}
    \item[] Justification: \rev{The paper does not present theorems. The derivation of the SFPO transition likelihood is given in Section~\ref{sec:planner_learning} and Appendix~\ref{sec:appendix_planner_sde}.}

    \item {\bf Experimental result reproducibility}
    \item[] Question: Does the paper fully disclose all the information needed to reproduce the main experimental results of the paper to the extent that it affects the main claims and/or conclusions of the paper (regardless of whether the code and data are provided or not)?
    \item[] Answer: \answerYes{}
    \item[] Justification: We fully describe our architecture settings, simulation and training parameters in \rev{Section~\ref{sec:Exp} and Appendices~\ref{app:experimental_protocol}, \ref{app:implementation_details}, and~\ref{app:training_details}}, and the provided information is sufficient to reproduce our main experimental results.

\item {\bf Open access to data and code}
    \item[] Question: Does the paper provide open access to the data and code, with sufficient instructions to faithfully reproduce the main experimental results, as described in supplemental material?
    \item[] Answer: \answerYes{}
    \item[] Justification: We provide sufficient material in supplemental material.

\item {\bf Experimental setting/details}
    \item[] Question: Does the paper specify all the training and test details (e.g., data splits, hyperparameters, how they were chosen, type of optimizer) necessary to understand the results?
    \item[] Answer: \answerYes{}
    \item[] Justification: We specify our training and test details in \rev{Section~\ref{metrics} and Appendices~\ref{app:experimental_protocol}, \ref{app:training_details}, and~\ref{app:algorithmic_derivations}.}

\item {\bf Experiment statistical significance}
    \item[] Question: Does the paper report error bars suitably and correctly defined or other appropriate information about the statistical significance of the experiments?
    \item[] Answer: \answerYes{}
    \item[] Justification: We define the evaluation metrics in \rev{Section~\ref{metrics}}. And then we compare our method with other baselines, showing statistical results in \rev{Sections~\ref{sec:results} and~\ref{sec:ablation}}.

\item {\bf Experiments compute resources}
    \item[] Question: For each experiment, does the paper provide sufficient information on the computer resources (type of compute workers, memory, time of execution) needed to reproduce the experiments?
    \item[] Answer: \answerYes{}
    \item[] Justification: Please see our experimental details in \rev{Section~\ref{sec:Exp} and Appendices~\ref{app:implementation_details} and~\ref{app:training_details}}. \rev{Training InfCoiL takes about one week on four NVIDIA GeForce RTX 4090 GPUs (Appendix~\ref{app:training_details}).}
    
\item {\bf Code of ethics}
    \item[] Question: Does the research conducted in the paper conform, in every respect, with the NeurIPS Code of Ethics \url{https://neurips.cc/public/EthicsGuidelines}?
    \item[] Answer: \answerYes{}
    \item[] Justification: We read the NeurIPS Code of Ethics, and make sure that our research conducted in the paper conform with the NeurIPS Code of Ethics in every respect.

\item {\bf Broader impacts}
    \item[] Question: Does the paper discuss both potential positive societal impacts and negative societal impacts of the work performed?
    \item[] Answer: \answerYes{}
    \item[] Justification: We discuss the societal impacts in \rev{Appendix~\ref{app:additional_discussion}}.
    
\item {\bf Safeguards}
    \item[] Question: Does the paper describe safeguards that have been put in place for responsible release of data or models that have a high risk for misuse (e.g., pre-trained language models, image generators, or scraped datasets)?
    \item[] Answer: \answerNA{}
    \item[] Justification: No such risks.

\item {\bf Licenses for existing assets}
    \item[] Question: Are the creators or original owners of assets (e.g., code, data, models), used in the paper, properly credited and are the license and terms of use explicitly mentioned and properly respected?
    \item[] Answer: \answerYes{}
    \item[] Justification: We cite works of related assets.

\item {\bf New assets}
    \item[] Question: Are new assets introduced in the paper well documented and is the documentation provided alongside the assets?
    \item[] Answer: \answerNA{}
    \item[] Justification: No new assets are introduced.

\item {\bf Crowdsourcing and research with human subjects}
    \item[] Question: For crowdsourcing experiments and research with human subjects, does the paper include the full text of instructions given to participants and screenshots, if applicable, as well as details about compensation (if any)? 
    \item[] Answer: \answerNA{}
    \item[] Justification: NA

\item {\bf Institutional review board (IRB) approvals or equivalent for research with human subjects}
    \item[] Question: Does the paper describe potential risks incurred by study participants, whether such risks were disclosed to the subjects, and whether Institutional Review Board (IRB) approvals (or an equivalent approval/review based on the requirements of your country or institution) were obtained?
    \item[] Answer: \answerNA{}
    \item[] Justification: NA

\item {\bf Declaration of LLM usage}
    \item[] Question: Does the paper describe the usage of LLMs if it is an important, original, or non-standard component of the core methods in this research? Note that if the LLM is used only for writing, editing, or formatting purposes and does \emph{not} impact the core methodology, scientific rigor, or originality of the research, declaration is not required.
    \item[] Answer: \answerNA{}
    \item[] Justification: NA

\end{enumerate}
\newpage

\appendix
\setlength{\parskip}{5.5pt plus 2pt}

\section{Experimental Protocol and Metrics}
\label{app:experimental_protocol}

This section defines the evaluation protocol, baseline variants, and metrics used in the main experiments.

\subsection{Evaluation Protocol and Baseline Variants}
\label{app:baselines}

All main comparisons are evaluated in the same Isaac Gym simulation environment with the same humanoid assets, object assets, and object physical parameters. We re-split OMOMO by partitioning the sequences of each morphology with the same train/test ratio, rather than using the original split that holds out two unseen morphologies for testing. We use a 30 Hz control rate by setting the simulation control-frequency divisor to two, and evaluate each method with deterministic controller actions. In the task-oriented protocol, each test sequence is evaluated three times and metrics are averaged. Planner training, controller training, and coordinated fine-tuning use only the training sequences. A rollout is successful at threshold $d$ (SR@$d$) if both the human pelvis and the object end within $d$ m of their target end positions. We use $d=0.3$ as the primary threshold and additionally report $d\in\{0.2, 0.1\}$ where available. In the reported tables, success rate is shown in percent, while humanoid tracking error $E_h$, object tracking error $E_o$, pelvis-goal error, and object-goal error are reported in centimeters. Table~\ref{tab:teacher_student_results} uses a separate InterMimic-style protocol, where success is termination-based rather than task-goal based.

\rev{\textbf{Baselines and variants.} The baselines and variants are described in Section~\ref{metrics}. InterMimic student uses the same observation normalization, action space, PD control, and evaluation code as our controller. InterMimic student, SimGenHOI*, and InfCoiL open-loop generate the complete reference once before execution, and the controller tracks it until success, failure, or timeout without feeding simulated states back to the planner. The closed-loop variants w/o FT P \& C, w/o FT C, and w/o FT P use the same receding-horizon execution as InfCoiL and differ only in which modules are fine-tuned.}

\textbf{InterMimic-style reference-tracking protocol.}
Table~\ref{tab:teacher_student_results} follows the InterMimic evaluation protocol rather than the task-oriented protocol in Table~\ref{tab:hoi_main_results}: rollouts are uniformly sampled from the dataset under a fixed budget, results are reported by the best sampled rollout, and SR means not triggering termination under the 0.5 m deviation threshold instead of reaching task goals.

\subsection{Metrics}
\label{app:metrics}

Success rate SR@$d$ is the fraction of evaluated episodes in which both the pelvis-goal and object-goal errors are below $d$ m, and SR without a threshold denotes SR@0.3. Avg.\ SR is the mean of SR@0.3, SR@0.2, and SR@0.1. \rev{Table~\ref{tab:matched_budget} reports all three thresholds for the closed-loop variants. For the open-loop methods, InterMimic student, SimGenHOI*, and InfCoiL open-loop reach 18.56\%, 45.12\%, and 43.58\% SR@0.2 and 9.52\%, 25.75\%, and 24.13\% SR@0.1 (Avg.\ SR 17.57\%, 41.00\%, and 38.98\%), all below the frozen closed-loop system at every threshold.} The pelvis-goal and object-goal errors are the final Euclidean distances to the corresponding target goals, reported in centimeters.

Humanoid tracking error $E_h$ is the mean Euclidean distance between selected simulated body keypoints and their reference keypoints, accumulated over the episode and reported in centimeters. Object tracking error $E_o$ is the mean Euclidean distance between sampled object surface points transformed by the simulated object pose and the same points transformed by the reference object pose, also reported in centimeters. These two metrics measure physical tracking quality rather than task completion alone: a method can reach the final goal but still have poor reference tracking, or track locally while failing to complete the object-centric task.

\section{Additional Experimental Results}
\label{app:additional_results}

This section reports supplementary quantitative and qualitative results that support the main experimental analysis.

\subsection{InterMimic-Style Reference-Tracking Evaluation}
\label{app:intermimic_tracking_eval}

Under the InterMimic-style protocol defined above, InfCoiL Teacher improves over InterMimic Teacher on both subsets, raising SR from 48.21\%/50.38\% to 92.86\%/93.56\% and reducing $E_h/E_o$ from 9.58/9.04 and 11.18/14.56 cm to 6.03/6.13 and 6.83/9.39 cm. The w/o CSI rows show that contact-aware state initialization is important for robust execution: removing CSI lowers SR by 25.00/7.58 points and worsens object tracking, despite a slightly lower FID on subset-8.

\begin{table}[ht]
\centering
\caption{InterMimic-style reference-tracking evaluation. Rollouts are uniformly sampled from the dataset under a fixed sampling budget and reported by the best sampled rollout. SR denotes not triggering termination under the 0.5 m deviation threshold, not task-goal success.}
\label{tab:teacher_student_results}

\small
\setlength{\tabcolsep}{10pt}
\begin{tabular}{@{}lccccccc@{}}
\toprule
\multirow{2}{*}{Method}
& \multirow{2}{*}{SR $\uparrow$}
& \multirow{2}{*}{$E_h \downarrow$ }
& \multirow{2}{*}{$E_o \downarrow$ }
& \multicolumn{3}{c}{Rprec $\uparrow$}
& \multirow{2}{*}{FID $\downarrow$} \\
\cmidrule(lr){5-7}
& & & & Top-1 & Top-2 & Top-3 & \\
\midrule
\multicolumn{8}{c}{subset-2 with ground-truth references} \\
\midrule
InterMimic Teacher 
& 48.21 & 9.58 & 9.04 & 12.50 & 31.25 & 34.38 & 10.36 \\
InfCoiL Teacher
& \textbf{92.86} & \textbf{6.03} & \textbf{6.13} & \textbf{18.75} & \textbf{43.75} & \textbf{53.12} & \textbf{3.29} \\
InfCoiL Teacher w/o CSI
& 67.86 & 7.01 & 13.34 & 9.38 & 28.12 & 43.75 & 3.98 \\

\midrule
\multicolumn{8}{c}{subset-8 with ground-truth references} \\
\midrule
InterMimic Teacher
& 50.38 & 11.18 & 14.56 & 19.53 & 35.16 & 46.88 & 5.88 \\
InfCoiL Teacher
& \textbf{93.56} & \textbf{6.83} & \textbf{9.39} & \textbf{23.83} & \textbf{42.19} & \textbf{55.08} & 1.76 \\
InfCoiL Teacher w/o CSI
& 85.98 & 7.97 & 14.92 & 21.48 & 38.67 & 53.12 & \textbf{1.46} \\

\bottomrule
\end{tabular}
\end{table}

\subsection{Relation to the InterMimic Evaluation Protocol}
\label{app:intermimic_gap}

The InterMimic student results in Table~\ref{tab:hoi_main_results} are lower than those reported by InterMimic~\cite{xu2025intermimic} because the two protocols measure different objectives, not because of a reproduction failure. We load the officially released student checkpoint with its official environment configuration and change only the references and the metric. InterMimic distills its student from successful teacher rollouts, omitting sequences that no teacher can track, and counts a rollout as successful if it tracks the reference without exceeding a 0.5 m deviation, reporting the best of repeated trials. Its success rate therefore measures reference survival on a teacher-supported subset. \rev{Moreover, each morphology-specific InterMimic teacher is trained on the full data of its subject before distillation, so the sequences in its reported OMOMO-Test evaluation are not held out from teacher training.} Our task-oriented SR instead requires both the pelvis and the object to reach their final goals on held-out test sequences, with planner-generated references and three averaged trials. Teacher filtering is also inapplicable to closed-loop generation. Future references depend on the states reached during execution, so filtering after execution would remove planning failures and reduce the task to post-hoc reference imitation. \rev{In summary, Table~\ref{tab:hoi_main_results} measures planner-conditioned task completion of the full system on held-out sequences, whereas the InterMimic-style evaluation in Table~\ref{tab:teacher_student_results} isolates controller tracking under ground-truth references.}

\subsection{Coordinated Learning at Matched Controller Budget}
\label{app:matched_budget}

\begin{table}[ht]
\caption{Coordinated learning at a matched controller budget under the closed-loop protocol of Table~\ref{tab:hoi_main_results}. C and P denote controller (PPO) and planner (SFPO) fine-tuning phases, with iteration counts in parentheses. Avg.\ SR averages the three thresholds, and errors are in cm.}
\label{tab:matched_budget}
\centering
\small
\setlength{\tabcolsep}{3pt}
\begin{tabular}{@{}lcccccccc@{}}
\toprule
Training path & SR@0.3$\uparrow$ & SR@0.2$\uparrow$ & SR@0.1$\uparrow$ & Avg.\ SR$\uparrow$ & $E_h\downarrow$ & $E_o\downarrow$ & Pelvis$\downarrow$ & Object$\downarrow$ \\
\midrule
Frozen closed loop & 56.09 & 48.91 & 30.59 & 45.20 & \textbf{7.25} & 16.10 & 46.89 & 69.83 \\
\midrule
C (10k) & 73.37 & 67.96 & 47.54 & 62.96 & 8.44 & 14.67 & 35.72 & 49.45 \\
C (10k) $\to$ P (500) & 77.24 & 72.07 & 48.59 & 65.97 & 8.90 & 14.76 & 32.90 & 45.37 \\
\midrule
C (14k) & 71.99 & 66.83 & 44.96 & 61.26 & 9.19 & \textbf{14.09} & 40.45 & 51.62 \\
C (10k) $\to$ P (500) $\to$ C (+4k) & \textbf{81.19} & \textbf{76.84} & \textbf{58.03} & \textbf{72.02} & 8.87 & 14.21 & \textbf{31.31} & \textbf{41.51} \\
\bottomrule
\end{tabular}
\end{table}

Table~\ref{tab:matched_budget} separates the contribution of planner learning from the number of controller updates. Starting from the frozen closed-loop system, 10k controller iterations raise SR@0.3 to 73.37\%, and a subsequent SFPO planner phase adds 3.87 points (77.24\%). Continuing controller-only training to 14k iterations does not help and lowers SR@0.3 to 71.99\%, whereas inserting the planner phase before the same additional 4k controller iterations reaches 81.19\%. At a matched controller budget of 14k iterations, coordinated learning therefore improves SR@0.3 by 9.20 points, Avg.\ SR by 10.76 points, and SR@0.1 by 13.07 points, and reduces pelvis/object goal errors from 40.45/51.62 cm to 31.31/41.51 cm. Planner adaptation thus raises the ceiling that controller adaptation alone reaches, and the gain of InfCoiL (C$\to$P$\to$C) does not come from additional controller updates.

\subsection{Online RL versus Supervised Fine-Tuning}
\label{app:sft_vs_rl}

\begin{table}[ht]
\caption{Planner update rules under closed- and open-loop execution. All rows use the controller after 10k closed-loop iterations (C (10k) in Table~\ref{tab:matched_budget}), and the planner is kept fixed, fine-tuned by SFT on executed trajectories as in SimGenHOI*, or fine-tuned online by SFPO. Open-loop rows execute references generated once by the same three planners.}
\label{tab:sft_vs_rl}
\centering
\small
\setlength{\tabcolsep}{6pt}
\begin{tabular}{@{}llcccc@{}}
\toprule
Execution & Planner update & SR@0.3$\uparrow$ & SR@0.2$\uparrow$ & SR@0.1$\uparrow$ & Avg.\ SR$\uparrow$ \\
\midrule
\multirow{3}{*}{Closed-loop} & None & 73.37 & 67.96 & 47.54 & 62.96 \\
 & SFT & 71.75 & 62.07 & 34.62 & 56.15 \\
 & Online RL (SFPO) & \textbf{77.24} & \textbf{72.07} & \textbf{48.59} & \textbf{65.97} \\
\midrule
\multirow{3}{*}{Open-loop} & None & 52.87 & 44.39 & \textbf{26.39} & 41.22 \\
 & SFT & \textbf{54.40} & 45.04 & 27.04 & \textbf{42.16} \\
 & Online RL (SFPO) & \rev{53.35} & \textbf{45.68} & 25.67 & \rev{41.57} \\
\bottomrule
\end{tabular}
\end{table}

Table~\ref{tab:sft_vs_rl} compares three planner update rules with the same controller-adapted checkpoint: no planner update, supervised fine-tuning (SFT) on trajectories executed by the controller, following the planner alignment of the SimGenHOI* baseline~\cite{lin2025simgenhoi}, and online RL with SFPO. For SFT, each training sequence is executed 10 times, and the successful execution with the lowest error is used as the fine-tuning target. The same three planners are also evaluated under open-loop execution, where each reference is generated once before execution. Under open-loop execution, SFT slightly improves over the fixed planner (52.87\% to 54.40\% SR@0.3 and 41.22\% to 42.16\% Avg.\ SR), and online RL changes the results only marginally (\rev{53.35}\% SR@0.3). Under closed-loop execution, however, SFT lowers SR@0.3 from 73.37\% to 71.75\% and Avg.\ SR from 62.96\% to 56.15\%, with the largest drop at the strictest threshold (47.54\% to 34.62\% SR@0.1). This is consistent with fitting near-distribution trajectories without a signal for recovering once replanning reaches states away from the demonstrations. Online RL instead assigns positive relative advantage to rollouts that recover and negative advantage to those that do not, and it is the only update that improves closed-loop performance (77.24\% SR@0.3 and 65.97\% Avg.\ SR).

\subsection{Planner RL Ablation: Setup}
\label{app:planner_rl_ablation}

This section provides the setup for the planner RL ablation in Sec.~\ref{sec:planner_rl_ablation}. We compare SFPO with FPO++~\cite{yi2026flow},\footnote{Our maximum sample size is limited to 8 instead of 16 or 32 because of memory constraints.} FPO++ with a behavior cloning anchor, ReinFlow~\cite{zhang2025reinflow} PPO with and without the anchor, and Flow-GRPO~\cite{liu2025flow} with isotropic noise. All methods are trained on a single NVIDIA 4090 GPU with a fixed learning rate of $2 \times 10^{-6}$. To ensure fair comparison and maintain the stability of the effective sample size for group-relative optimization, ReinFlow and the GRPO-based methods, which share the same noise injection scheme, use a batch size of 1536 environments. In contrast, FPO++ consumes significantly more memory and is restricted to a batch size of 512. Since FPO++ and ReinFlow require a critic network for value estimation, they share the same critic network architecture. Detailed implementation and hyperparameter settings are provided in Appendix~\ref{sec:appendix_fpo_bc_critic}.

Since FPO++ does not use denoising-path noise injection, its BC anchor is implemented as an output-space distance to a frozen base planner. Appendix~\ref{sec:appendix_bc_anchor} defines this distance and the frozen baseline.

\subsection{Planner Prediction and Loss Parameterization}
\label{app:xv_ablation}

We compare our planner, which predicts the clean state $x_0$ and is trained with an $x$-space loss (X-base), against a standard rectified-flow planner that predicts the velocity $v$ and is trained with a velocity-space loss (V-base). The two variants differ only in this parameterization and share the same architecture, training data, and number of training epochs. We evaluate open-loop autoregressive generation on the official OMOMO test split against the ground-truth motions, following the metric definitions of OMOMO~\cite{li2023object}: foot sliding (FS), mean per-joint position error (MPJPE), root and object translation errors ($T_{\mathrm{root}}$, $T_{\mathrm{obj}}$), object orientation error ($O_{\mathrm{obj}}$), and mean per-joint velocity error (MPJVE).

\begin{table}[ht]
\centering
\caption{Planner prediction and loss parameterization, evaluated by open-loop autoregressive generation on the official OMOMO test split. V-base predicts $v$ with a velocity-space loss, and X-base predicts $x_0$ with an $x$-space loss.}
\label{tab:xv_ablation}
\small
\begin{tabular}{@{}lcccccc@{}}
\toprule
Variant & FS$\downarrow$ & MPJPE$\downarrow$ & $T_{\mathrm{root}}\downarrow$ & $T_{\mathrm{obj}}\downarrow$ & $O_{\mathrm{obj}}\downarrow$ & MPJVE$\downarrow$ \\
\midrule
V-base & 0.579 & \textbf{11.71} & \textbf{7.40} & 17.25 & 1.098 & 1.122 \\
X-base (ours) & \textbf{0.350} & 11.87 & 7.43 & \textbf{16.06} & \textbf{1.046} & \textbf{0.686} \\
\bottomrule
\end{tabular}
\end{table}

Table~\ref{tab:xv_ablation} shows that X-base reduces foot sliding by 40\% and velocity error by 39\% relative to V-base and lowers object translation and orientation errors, while V-base is marginally better on MPJPE (11.71 vs.\ 11.87) and root translation (7.40 vs.\ 7.43). Because the controller must track these references under contact-rich dynamics, we adopt X-base for InfCoiL.

\subsection{Generalization to Random Morphology--Object Combinations}

\begin{figure}[t]
  \centering
  \includegraphics[width=\linewidth,trim=0 51 0 0,clip]{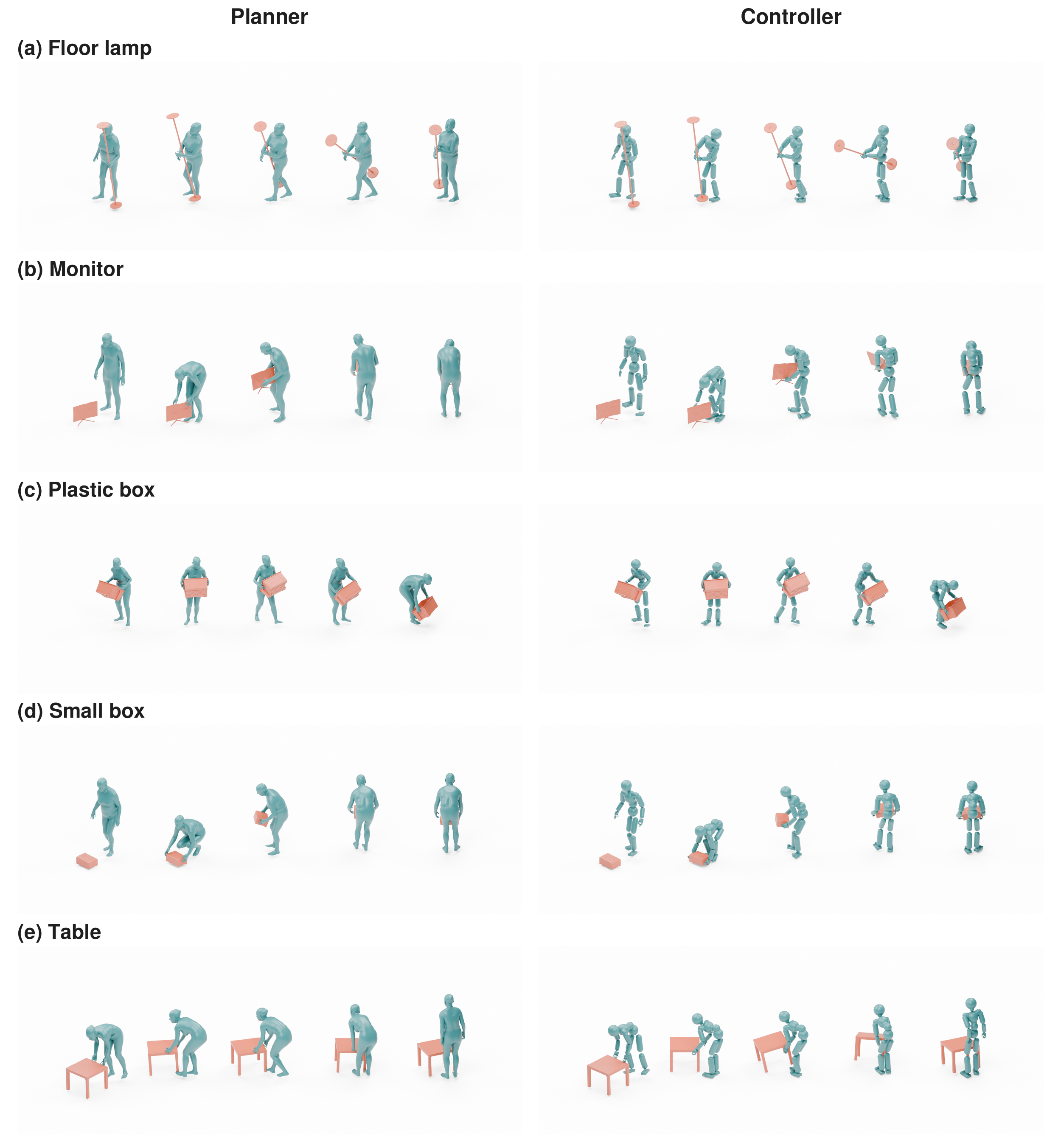}
  \caption{Planner references (left) and controller motions (right) for floorlamp, monitor, plasticbox, smallbox, and table. Each row shows matching frames from one episode, ordered from left to right.}
  \label{fig:random_sampled_beta_obj}
\end{figure}

\begin{figure}[t]
  \centering
  \includegraphics[width=\linewidth,trim=0 66 0 0,clip]{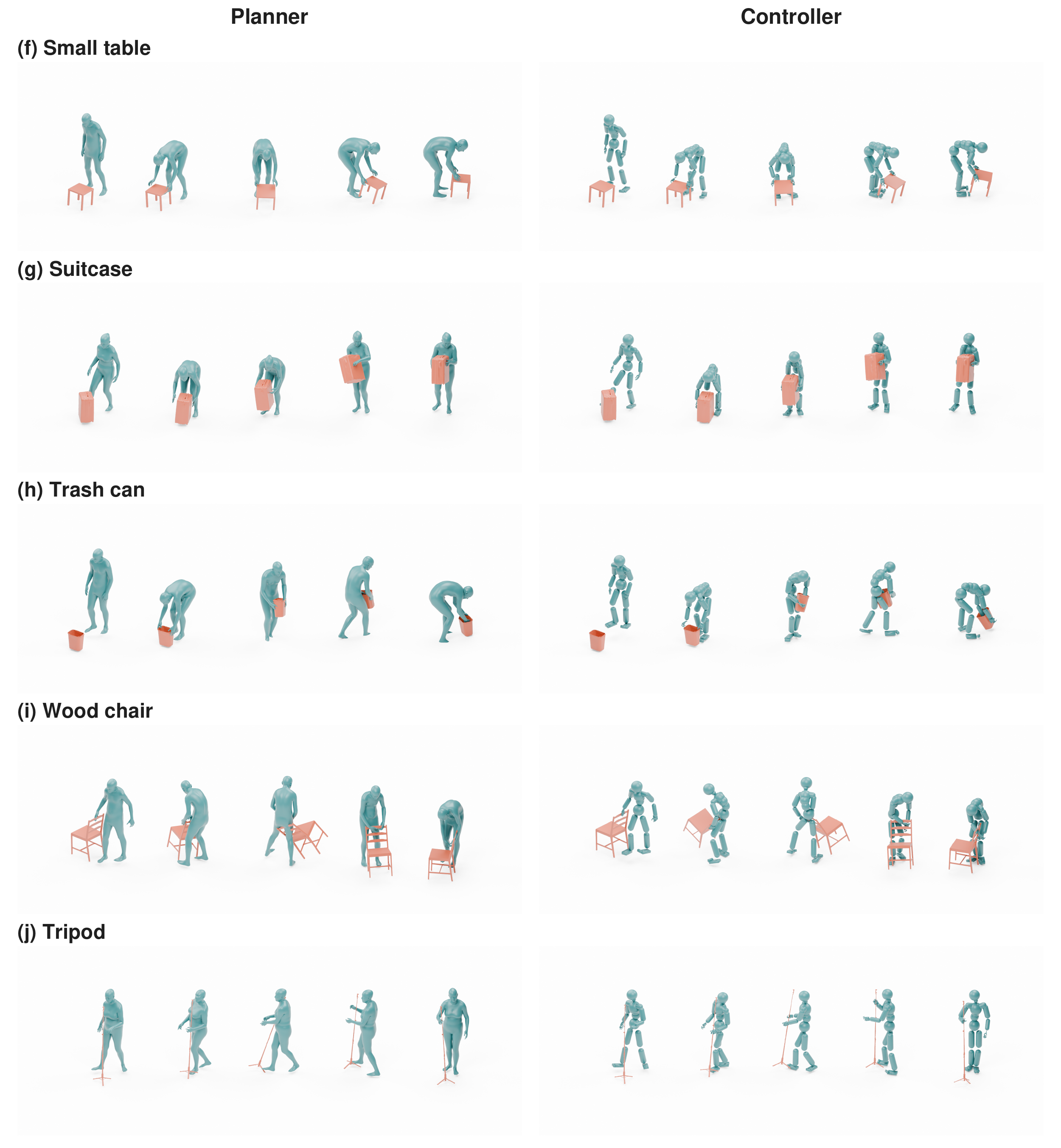}
  \caption{Planner references (left) and controller motions (right) for smalltable, suitcase, trashcan, woodchair, and tripod. Each row shows matching frames from one episode, ordered from left to right.}
  \label{fig:random_sampled_beta_obj_part2}
\end{figure}

Figures~\ref{fig:random_sampled_beta_obj} and~\ref{fig:random_sampled_beta_obj_part2} compare planner references and controller motions from the same episodes across different body shapes and objects. Across diverse objects such as floorlamp, plasticbox, smalltable, and trashcan, the planner produces semantically meaningful SMPL-X interaction references, and the humanoid controller tracks these references in simulation. Although the controller embodiment differs from the planner morphology and may adjust poses, contact timing, and object trajectories during execution, it largely preserves the intended interaction semantics, such as bending toward the object, lifting, carrying, or repositioning it. These results demonstrate that the closed-loop planner-controller design improves robustness to morphology-object variation and supports generalization beyond fixed body-object pairs.

\label{app:random_sampled_beta_obj_cropped}

\subsection{\texorpdfstring{\rev{Qualitative Failure Analysis}}{Qualitative Failure Analysis}}
\label{app:failure_analysis}

\rev{\textbf{Failure modes of open-loop execution.} Figure~\ref{fig:expcomp} shows typical failures of the open-loop baselines. In the suitcase example, InterMimic student tips the suitcase over while approaching it and never establishes a lifting grasp, whereas SimGenHOI* lifts the suitcase but carries it in a tilted pose and does not complete the task. In the monitor example, both baselines lose control of the object after contact, and SimGenHOI* drops it to the ground. In the small-box example, InterMimic student briefly lifts the box but loses its grasp, and the rollout is recorded as a failure. These examples illustrate a limitation of open-loop execution. The reference is generated once before execution; after a deviation in human-object contact, the controller continues tracking the fixed reference, and errors can accumulate until termination.}

\rev{\textbf{Effect of closed-loop replanning and coordinated learning.} InfCoiL completes the same tasks because the planner replans every $N_r$ steps from the realized human, object, and contact states, so each new reference window starts from the actual object configuration rather than the one assumed before execution. Table~\ref{tab:hoi_main_results} quantifies this effect. With the same frozen planner and controller, closed-loop replanning raises SR from 49.23\% to 56.09\% and reduces pelvis/object goal errors from 57.15/75.51 cm to 46.89/69.83 cm. The remaining gap suggests that replanning alone cannot compensate for references that the controller fails to execute precisely around contact. Controller adaptation raises SR to 73.37\%, and subsequent planner learning moves references toward behaviors that the adapted controller executes reliably, reaching 77.24\% (C$\to$P) and 81.19\% (C$\to$P$\to$C).}

\rev{\textbf{Remaining failures.} Inspecting failed InfCoiL rollouts, we observe two recurring regimes. First, objects with complex contact geometry or less reliable simulated dynamics, such as the monitor, remain harder to grasp and manipulate stably than simple boxes. Second, when an object is initialized in an airborne or otherwise strongly off-demonstration state, the planner may not react before termination, because the demonstrations contain little or no pre-contact object motion. Neither regime is resolved by replanning or coordinated learning on the current data, and addressing them would likely require broader coverage of object states in the training data and more accurate contact modeling.}

\section{System Implementation Details}
\label{app:implementation_details}
\label{app:system_implementation}

This section describes the concrete planner-controller interface, planner tokenization and conditioning, controller observation design, and action parameterization needed to reproduce InfCoiL.

\subsection{Planner Representation and Conditioning}
\label{app:planner_details}

The high-level planner predicts structured HOI motion tokens over a short horizon. Each frame token has 490 channels:
\begin{equation}
490 = 72_{\mathrm{body}} + 90_{\mathrm{hand}} + 312_{\mathrm{rot}} + 3_{\mathrm{obj\ trans}} + 9_{\mathrm{obj\ rot}} + 4_{\mathrm{contact}} .
\end{equation}
The first term contains 24 body joints in 3D, the second contains 30 hand joints in 3D, the rotation term stores 52 global joint rotations in 6D representation, the object state contains object translation and a $3\times3$ rotation matrix, and the final term stores coarse hand-object contact labels. During closed-loop execution, the planner receives the latest simulated humanoid, object, and contact history, converts it back into this token format, and clamps the prefix frames during sampling. This makes the generated future window continuous with the physical state actually reached by the controller.

The planner uses a Transformer encoder with model width 512, 16 attention heads, 8 layers, GELU activations, dropout 0.1, and sinusoidal positional encoding. To keep each token compact, the 490-dimensional frame feature is split into two motion sub-tokens of 220 and 270 dimensions, embedded separately, interleaved over time, and then decoded back to the original channel order. We train the model as a rectified-flow planner with predict-$x$ and loss-$x$ parameterization, using logit-normal time sampling, Gaussian noise scale 1, and 10 Euler sampling steps at inference. The supervised planner is trained with batch size 512, learning rate $10^{-4}$, EMA decay 0.9998, and auxiliary losses on object keypoints, forward-kinematic hand/foot positions, and finger-tip local offsets.

Conditioning is encoded as prefix tokens before the motion tokens. Following the OMOMO-based language-conditioned HOI setting of CHOIS~\cite{li2024controllable}, the language instruction is encoded by a pretrained CLIP ViT-L/14@336px text encoder~\cite{radford2021learning}, which is kept frozen in evaluation mode rather than trained from scratch. Its L2-normalized 768-dimensional sentence embedding is projected to the planner width by a trainable two-layer MLP and fused with a dynamic progress embedding computed from the current progress index, the target end index, and the sequence length. The pelvis goal is encoded from the local horizontal $(x,z)$ coordinates, while the object goal is encoded from its 3D local position. Object geometry is represented by BPS features using 1024 object surface points and projected by a two-layer MLP into the planner width. Morphology is encoded from the 16-dimensional SMPL-X shape vector $\beta$ through a linear projection. During supervised training, the morphology token is randomly dropped with probability 0.1 for classifier-free-style regularization, and normal inference uses the conditional branch. Since the text embedding is part of the conditioning context $c$, it affects the predicted clean state and its induced velocity at every flow step. The interface accepts free-form sentences that the CLIP tokenizer can process, rather than a finite set of prompt IDs, without architectural changes. Its effective semantic range, however, follows the interaction types and phrasing of the OMOMO/CHOIS annotations, since no additional language corpus is used. Deterministic evaluation uses the selected annotation of each test sequence.

The planner operates in a receding-horizon manner. The HOI model samples a 16-token future window with 4 additional prefix frames, and the sampled trajectory is converted to a 30 Hz simulation reference by interpolation. In our default closed-loop setting, this yields an executable reference horizon of $N_f=42$ frames, while replanning is triggered every $N_r=24$ simulation steps. Since the horizon is longer than the replanning interval, consecutive plans overlap, allowing the controller to track a near-future reference while the planner periodically refreshes later motion from updated physical feedback.

\subsection{Controller Observations, Architecture, and Actions}
\label{app:controller_details}

The controller uses a unified student policy for all SMPL-X body shapes. A single per-step tracking observation has 1599 dimensions and contains heading-normalized humanoid proprioception, contact indicators, object state, reference-state differences, future reference features, and interaction-geometry features between selected body keypoints and object surface points. The student policy receives four temporal versions of this tracking observation at offsets $\{0,1,4,16\}$, concatenated with the 16-dimensional SMPL-X shape vector:
\begin{equation}
o^{\mathrm{student}}_t =
\left[\beta,\ o_t^{0},\ o_t^{1},\ o_t^{4},\ o_t^{16}\right] \in \mathbb{R}^{6412}.
\end{equation}
This observation design gives the controller both the current tracking error and near-future reference context, while the shape vector tells the shared policy which morphology is being controlled.

The student actor is a compact Transformer encoder. The shape vector is embedded into a 256-dimensional shape token, and each of the four 1599-dimensional observation vectors is embedded into a 256-dimensional observation token. The five tokens are summed with sinusoidal positional encoding and processed by a 3-layer Transformer encoder with 4 attention heads, 512 feed-forward width, GELU activations, and zero dropout. The output corresponding to the shape token is used as a CLS-style summary and projected to the Gaussian action mean. The critic follows the standard actor-critic implementation used by the underlying RL framework and predicts the value from the normalized full observation.

The action is a 153-dimensional vector corresponding to the actuated humanoid joint degrees of freedom. The simulated SMPL-X humanoid has 52 rigid bodies. Excluding the floating root, the controller actuates 51 joints with a 3D local rotation target per joint. Actions are clipped to $[-1,1]$ and converted to PD position targets by
\begin{equation}
q^{\mathrm{target}} = q^{\mathrm{offset}} + q^{\mathrm{scale}} \odot a ,
\end{equation}
where $q^{\mathrm{offset}}$ and $q^{\mathrm{scale}}$ are computed from joint limits. The simulator then applies position-control drives to track these targets. Thus, the policy does not directly output torques but normalized joint-position targets for a low-level PD controller.

\section{Training Details}
\label{app:training_details}

This section collects controller distillation, contact-aware initialization, closed-loop fine-tuning, and rollout settings for coordinated planner-controller learning. \rev{Training InfCoiL takes about one week on four NVIDIA GeForce RTX 4090 GPUs.}

\subsection{Controller Distillation and Fine-Tuning}
\label{app:controller_training}

The unified controller is initialized by distilling subject-specific InterMimic-style teachers into one multi-morphology student with the distillation-to-RL curriculum of InterMimic~\cite{xu2025intermimic}. During rollout, each environment is assigned a subject-object pair. At every simulation step and independently for each environment, the executed action is the corresponding teacher's sampled and clipped action with probability \rev{$p_{\mathrm{tea}}(e)$} and the student's action otherwise, where $\rev{p_{\mathrm{tea}}(e)}=\operatorname{clip}\left(1-(e-500)/5000,\,0,\,1\right)$ and $e$ is the training epoch. Rollouts are thus fully teacher-driven for the first 500 epochs and fully student-driven after epoch 5500. Regardless of which policy acts, the teacher's action mean on the visited state is recorded as the imitation target for the student's action mean, providing a DAgger-style signal on states increasingly induced by the student.

The training objective follows a gated schedule. For $e \le 5000$, only the imitation loss is used. For $5000 < e \le 6000$, the critic loss is added with a weight increasing linearly to its full value, after which it remains at full weight. From $e > 7000$, we track an exponential moving average of the critic's explained variance, updated as $0.99\,\mathrm{EV} + 0.01\,\mathrm{ev}$ at every minibatch. The PPO actor loss is applied to a minibatch only when this average has reached 0.6 for three consecutive minibatch checks. Otherwise, the minibatch falls back to the critic and full-weight imitation losses. Letting $n$ be the cumulative number of gated minibatch updates, the actor and imitation weights are $\min(n/4000,1)$ and $\max(1-n/4000,0.1)$, respectively.

We use PPO with discount factor 0.99, GAE parameter 0.95, learning rate $2\times10^{-5}$, clipping threshold 0.2, horizon length 16, minibatch size 16384, 6 mini-epochs, critic coefficient 5, gradient-norm limit 1, and action-bound loss coefficient 10. The Gaussian exploration parameter is fixed during student training. The base imitation coefficient is 1.0 and is annealed to 0.1 within the same pretraining run as described above. In closed-loop controller fine-tuning it is set to 0 so that the controller is optimized directly for the closed-loop reward induced by planner-generated references.

For contact-rich phases, we use contact-aware state initialization during controller pretraining. Demonstration frames near hand-object contact onset and release receive higher reset probability, controlled by a contact window of 15 frames and sampling weight scale 3.0. This exposes the student to difficult grasping and release states more often than uniform random initialization, reducing the chance that rare but important contact transitions are under-trained.

\subsection{Closed-Loop Training Details}
\label{app:closed_loop_training}

Closed-loop controller fine-tuning uses the same simulator, action parameterization, and student architecture as pretraining, but the reference trajectory is generated online by the planner. In the default all-subject configuration, the environment uses 4096 parallel simulations, start-state initialization, 500-step episode length, 24-step replanning interval, and 42-frame reference horizon. The reward keeps humanoid and object tracking terms active, adds terminal success and termination rewards, and disables the interaction-graph and contact-consistency terms during closed-loop coordinated learning.

For planner fine-tuning, SFPO evaluates a sampled plan only after execution by the controller and physics simulator. Rollouts are grouped by identical task semantics, including subject, object, sequence, reset state, start time, and goal conditions, so samples in the same group differ only by the stochastic flow transitions used for planner exploration. We compute a sequence-level score from the time-averaged transition reward plus terminal task reward, normalize it within the group, clip the resulting advantage, and assign the same sequence advantage to all planner transitions produced during that rollout. This avoids comparing absolute rewards across tasks with different objects, body shapes, or reset difficulties.

\section{Planner RL and Algorithmic Derivations}
\label{sec:appendix_method_details}
\label{app:algorithmic_derivations}

This section provides the planner-side derivations and optimization details omitted from the main paper, including SFPO, FPO baselines, group-relative advantages, and the coordinated learning schedule.

\subsection{Derivations for SFPO}
\label{sec:appendix_planner_sde}

\rev{This section derives the SFPO transition likelihood of Section~\ref{sec:planner_learning}.} The planner update follows the ReinFlow/Flow-GRPO stochastic-denoising view: selected denoising transitions are treated as Gaussian policy steps, the old transition log-probabilities are stored during rollout, and the current log-probabilities are recomputed on the same denoising path during PPO-style updates. Our modification is semantic: the transition variance is diagonal and channel-dependent, reflecting that body translations, rotations, hand joints, object states, and contact labels tolerate different perturbation scales. This keeps the online RL path aligned with Flow-GRPO's stochastic flow policy while avoiding an explicit density-score estimator for HOI motion tokens.

Consider a $K$-step Euler denoising sampler. Let
\begin{equation}
z_k = x_{\tau_k}, \quad k = 1, \dots, K+1,
\end{equation}
where $\tau_k \in [0, 1]$ denotes the flow matching time at the $k$-th denoising step, with $\tau_1 = 1$ and $\tau_{K+1} = 0$. Stochastic transitions are applied to a fixed subset of $S$ denoising steps,
\begin{equation}
\mathcal{S} \subseteq \{1, \dots, K\}, \quad |\mathcal{S}| = S.
\end{equation}
For steps outside this subset ($k \notin \mathcal{S}$), the planner retains the original deterministic updates. For the selected steps ($k \in \mathcal{S}$), we model the transition as a Gaussian policy. Selecting only a subset of denoising steps sacrifices the accuracy of the full-path log-probability but significantly reduces computation and memory overhead. When $S$ equals $K$, this formulation degenerates to a full-chain stochastic flow policy. In all experiments, we use $K=10$ and make only the final step stochastic, \rev{\emph{i.e.}}, $\mathcal{S}=\{K\}$ and $S=1$, corresponding to the transition from $\tau_K=0.1$ to $\tau_{K+1}=0$. This subset is fixed across all rollouts rather than resampled, and the preceding $K-1$ steps use deterministic Euler updates.

For the selected denoising steps, we start from the marginal-preserving reverse-time SDE used by Flow-GRPO:
\begin{equation}
d x_\tau =
\left[
v_\theta(x_\tau,\tau,c_t)
- \frac{1}{2}\sigma_\tau^2 \nabla_{x_\tau}\log p_\tau(x_\tau \mid c_t)
\right]d\tau
+ \sigma_\tau dW_\tau,
\label{eq:appendix_flowgrpo_score_sde}
\end{equation}
where $v_\theta$ denotes the velocity field of the planner, $c_t$ is the condition at \rev{simulation step} $t$, and $\sigma_\tau$ controls the stochasticity. For rectified flow with linear interpolation, the score can be rewritten as
\begin{equation}
\nabla_{x_\tau}\log p_\tau(x_\tau \mid c_t)
= -\frac{x_\tau + (1-\tau)v_\theta(x_\tau,\tau,c_t)}{\tau}.
\end{equation}
Substitution yields a score-free drift expressed only through the current state and velocity:
\begin{equation}
\bar{v}_\theta(x_\tau,\tau,c_t)
= v_\theta(x_\tau,\tau,c_t)
+ \frac{\sigma_\tau^2}{2\tau}
\left[x_\tau + (1-\tau)v_\theta(x_\tau,\tau,c_t)\right].
\label{eq:appendix_score_free_drift}
\end{equation}
We use this score-free rectified-flow form rather than estimating $\nabla\log p_\tau$ with an additional model. For HOI tokens, an explicit score estimator would need to model a mixed representation containing normalized continuous body/object coordinates, rotations, hand articulation, autoregressive fixed prefixes, and contact-like semantic channels. Errors in such a score are particularly harmful because they enter the drift at every stochastic denoising step and can move samples into contact-inconsistent states before the physics controller evaluates them. The velocity-substituted form retains Flow-GRPO's Gaussian transition likelihood while keeping the planner architecture unchanged.

We further replace the isotropic diffusion coefficient with a semantic diagonal coefficient:
\begin{equation}
\Gamma_\tau = \lambda_{\mathrm{n}} \sqrt{\frac{\tau}{1-\tau}} \operatorname{Diag}(\rev{\kappa}),
\label{eq:appendix_semantic_gamma}
\end{equation}
where $\lambda_{\mathrm{n}}$ denotes the global noise level and $\rev{\kappa} \in \mathbb{R}_+^D$ assigns a fixed scale to each channel, so that channel $j$ has standard deviation proportional to \rev{$\kappa_j$}. The scales are set by hand, constant within each of six channel groups, and not updated during RL. Table~\ref{tab:sfpo_q} lists the scale of each channel group. Object translation receives the largest perturbation and rotations the smallest. The noise levels compared in Table~\ref{tab:planner_rl} ($\lambda_{\mathrm{n}}\in\{0.05,0.1,0.2\}$) change only the global level. \rev{In the discrete sampler, the time factor is computed as $\sqrt{\tau_k/\max(1-\tilde\tau_k,10^{-6})}$, with $\tilde\tau_k=\tau_{k+1}$ at the endpoint $\tau_k=1$ and $\tilde\tau_k=\tau_k$ otherwise}, which does not affect our experiments, where the only stochastic step has $\tau_K = 0.1$. The isotropic Flow-GRPO baseline in Table~\ref{tab:planner_rl} uses the same time schedule with identical entries of \rev{$\kappa$} across all perturbable channels. With this diagonal covariance, the score-free drift in \rev{Eq.~\eqref{eq:appendix_score_free_drift}} is applied elementwise as
\begin{table}[ht]
\centering
\caption{Semantic noise scales \rev{$\kappa$} used by SFPO, constant within each channel group of the 490-dimensional planner frame.}
\label{tab:sfpo_q}
\small
\begin{tabular}{@{}lcc@{}}
\toprule
Channel group & Dim. & \rev{$\kappa$} \\
\midrule
Body joint positions & 72 & 0.35 \\
Hand joints & 90 & 0.175 \\
Human global rotations & 312 & 0.07 \\
Object translation & 3 & 0.7 \\
Object rotation & 9 & 0.07 \\
Contact labels & 4 & 0.175 \\
\bottomrule
\end{tabular}
\end{table}
\begin{equation}
\bar{v}_\theta(z_k,\tau_k,c_t)
= v_\theta(z_k,\tau_k,c_t)
+ \frac{1}{2\tau_k}\Gamma_{\tau_k}\Gamma_{\tau_k}^{\top}
\left[z_k + (1-\tau_k)v_\theta(z_k,\tau_k,c_t)\right].
\end{equation}
Through Euler-Maruyama discretization, the transition for the selected steps becomes
\begin{equation}
z_{k+1} = \mu_{\theta,k} + \Gamma_{\tau_k}\sqrt{-\Delta\tau_k}\epsilon_k, \quad \epsilon_k \sim \mathcal{N}(0, I),
\label{eq:sfpo_noise}
\end{equation}
where the mean is given by
\begin{equation}
\mu_{\theta,k} = z_k + \bar{v}_\theta(z_k, \tau_k, c_t)\Delta\tau_k, \quad \Delta\tau_k = \tau_{k+1} - \tau_k < 0.
\end{equation}
Therefore, the policy probability is expressed as
\begin{equation}
\pi_\theta(z_{k+1} \mid z_k, c_t) = \mathcal{N} \left( \mu_{\theta,k}, (-\Delta\tau_k)\Gamma_{\tau_k}\Gamma_{\tau_k}^{\top} \right).
\end{equation}

During the planner rollout phase, the frozen snapshot $\theta_{\mathrm{old}}$ generates plans and we record their transition log-probabilities. In the update phase, we recompute them on the same denoising path with the current planner. Each replanning event $t$ of the $g$-th rollout under task condition $c_i$ forms one training sample\rev{, whose context $c_t$ combines $c_i$ with the executed history}. Let \rev{$\Omega$} denote the generated elements \rev{$(u,d)$} of a planned window, \emph{i.e.}, its non-prefix frames \rev{$u$} times the $D=490$ channels \rev{$d$}. We use a normalized log-probability score, in which the initial-noise term $z_1\sim\mathcal{N}(0,I)$ and each selected transition term are computed as the mean element-wise log-density over the same set \rev{$\Omega$}, and these $S+1$ terms are then averaged:
\begin{equation}
\bar\ell_\theta^{\,i,g,t}=\operatorname{clip}\Big(\frac{1}{S+1}\Big[\frac{1}{|\rev{\Omega}|}\sum_{\rev{(u,d)\in\Omega}}\log\mathcal{N}(z_{1,\rev{u,d}};0,1)+\sum_{k\in\mathcal{S}}\frac{1}{|\rev{\Omega}|}\sum_{\rev{(u,d)\in\Omega}}\log\pi_{\theta,\rev{u,d}}(z_{k+1,\rev{u,d}}\mid z_k,c_t)\Big],\,-10,\,10\Big),
\label{eq:appendix_logprob}
\end{equation}
where $\pi_{\theta,\rev{u,d}}$ is the univariate marginal of Eq.~\eqref{eq:sfpo_policy} for element \rev{$(u,d)$}, and all states are those of sample $(i,g,t)$. The importance ratio is
\begin{equation}
\rho_\theta^{i,g,t}=\exp\big(\bar\ell_\theta^{\,i,g,t}-\bar\ell_{\theta_{\mathrm{old}}}^{\,i,g,t}\big).
\end{equation}
The initial-noise term does not depend on $\theta$ and cancels in the difference when clipping is inactive, but it halves the scale of the log-ratio for $S=1$. Together with the element-wise averaging, this makes $\rho$ a normalized surrogate ratio rather than a ratio of joint Gaussian densities over the planned window, and the ratio clipping range $\epsilon_{\mathrm{c}}$ in Table~\ref{tab:sfpo_hparams} is set on this scale.

Sampling $G$ rollouts under the same task condition $c_i$ yields sequence-level rewards $R_i^1, \dots, R_i^G$ after execution by the controller and the physics engine, and every replanning event of rollout $g$ receives the clipped group-relative advantage $A_i^g$ of Eq.~\eqref{eq:clipped_advantage}. Appendix~\ref{sec:appendix_group_advantage} describes how groups are formed and how invalid rollouts and zero-variance groups are excluded. The planner objective is
\begin{equation}
\begin{aligned}
\rev{\mathcal{J}_{\mathrm{clip}}}&=\mathbb{E}_{i,g,t}\Big[\min\big(\rho_\theta^{i,g,t}A_i^g,\ \operatorname{clip}(\rho_\theta^{i,g,t},1-\epsilon_{\mathrm{c}},1+\epsilon_{\mathrm{c}})A_i^g\big)\Big],\\
\mathcal{J}_{\mathrm{SFPO}}&=\rev{\mathcal{J}_{\mathrm{clip}}-\lambda_{\mathrm{KL}}\,\mathbb{E}_{i,g,t,\,k\in\mathcal{S}}\Big[D_{\mathrm{KL}}\big(\pi_\theta(\cdot\mid z_k,c_t)\,\Vert\,\pi_{\mathrm{fix}}(\cdot\mid z_k,c_t)\big)\Big]}.
\end{aligned}
\end{equation}
The expectation runs over all valid replanning samples, so longer rollouts contribute more samples. Ratios are neither multiplied across the replanning events of a rollout nor normalized by their number. In practice, the loss is first averaged over valid samples within each log-probability microbatch and then weighted by the microbatch size, which differs slightly from the global valid-sample mean when a microbatch contains invalid samples. No fixed-reference KL or BC penalty is applied in the reported SFPO configuration (\rev{$\lambda_{\mathrm{KL}}=0$}). Table~\ref{tab:sfpo_hparams} lists the SFPO hyperparameters.

\begin{table}[ht]
\centering
\caption{SFPO hyperparameters used in all reported planner fine-tuning phases.}
\label{tab:sfpo_hparams}
\small
\begin{tabular}{@{}ll@{}}
\toprule
Hyperparameter & Value \\
\midrule
Group size $G$ & 16 \\
Denoising steps $K$ / stochastic steps $S$ & 10 / 1 (final step, $\tau$: $0.1\to0$) \\
Global noise level $\lambda_{\mathrm{n}}$ & 0.1 \\
Ratio clipping $\epsilon_{\mathrm{c}}$ & 0.03 \\
Advantage clipping $A_{\max}$ & 5 \\
Group skipped if reward std $\le$ & $10^{-6}$ \\
Std floor in advantage normalization & $10^{-12}$ \\
Log-probability clipping & $[-10, 10]$ \\
Reference regularizer weight \rev{$\lambda_{\mathrm{KL}}$} & 0 \\
PPO epochs per iteration & up to 5, \rev{early stop when approx.\ KL exceeds $0.01$} \\
Optimizer minibatch & 50{,}000 replanning samples \\
Log-probability microbatch & 1{,}024 samples \\
Planner sampling chunk & 128 \\
Learning rate & $2\times10^{-6}$ \\
\bottomrule
\end{tabular}
\end{table}

\subsection{FPO Baseline, BC Anchor, and Critic Details}
\label{sec:appendix_fpo_bc_critic}

The FPO ablations reported in the experiments are distinct from the SFPO planner update above. They reuse the same closed-loop planner-controller rollout and \rev{the same reward and scoring pipeline}, but the actor ratio is not computed from ReinFlow denoising-path log-probabilities. Instead, FPO constructs a conditional-flow-matching surrogate ratio on the final denoised plan.

\subsubsection{FPO Ratio}
\label{sec:appendix_fpo_ratio}

During an FPO rollout, the planner first generates a final denoised motion window $x_0$ using \rev{the planner's flow-matching sampler}. For each generated window, FPO samples auxiliary pairs $(\epsilon_s,\tau_s)$\rev{, $s=1,\dots,N_{\mathrm{aux}}$}, with $\epsilon_s \sim \mathcal{N}(0,I)$ and $\tau_s \in (0,1)$, and interpolates
\begin{equation}
x_{\tau_s} = \tau_s \epsilon_s + (1-\tau_s)x_0 .
\end{equation}
The autoregressive fixed-prefix frames are clamped before the model forward pass. Since \rev{our planner} predicts $x_0$, \rev{FPO uses the induced velocity of Eq.~\eqref{eq:velocity}},
\begin{equation}
\rev{v_\theta}(x_{\tau_s},\tau_s,c)
= \frac{x_{\tau_s}-\rev{f_\theta}(x_{\tau_s},\tau_s,c)}{\tau_s},
\end{equation}
and the target velocity is $v^\star=\epsilon_s-x_0$. FPO computes the masked CFM loss
\begin{equation}
\rev{\mathcal{L}^{\mathrm{CFM}}_\theta}(x_0,\epsilon_s,\tau_s,c)
= \frac{1}{|\Omega|}\sum_{(u,d)\in\Omega}
\rev{\big(v_{\theta,u,d} - v^\star_{u,d}\big)^2},
\end{equation}
where $\Omega$ \rev{is the generated-element set of Eq.~\eqref{eq:appendix_logprob}, which} excludes the fixed autoregressive prefix. The rollout stores \rev{$\mathcal{L}^{\mathrm{CFM}}_{\mathrm{old}}$} from the \rev{rollout model $\theta_{\mathrm{old}}$}. During training, the current model recomputes \rev{$\mathcal{L}^{\mathrm{CFM}}_\theta$} on the same $(x_0,\epsilon_s,\tau_s,c)$ samples and forms
\begin{equation}
\rho_\theta^{\mathrm{FPO}}
= \exp\left(\rev{\mathcal{L}^{\mathrm{CFM}}_{\mathrm{old}}}-\rev{\mathcal{L}^{\mathrm{CFM}}_\theta}\right).
\end{equation}
The actor loss then applies the same proximal trust-region principle as PPO, with clipping to prevent extreme surrogate-ratio updates. This FPO ratio should therefore be interpreted as a CFM-loss surrogate baseline, not as the denoising-transition likelihood used by SFPO.

\subsubsection{BC Anchor}
\label{sec:appendix_bc_anchor}

The BC Anchor constrains FPO updates to remain close to a frozen motion prior in output space. The baseline model is initialized from an explicit checkpoint if provided. Otherwise, it is copied from the fixed base reference when that exists, or from the initial \rev{planner} parameters. Its parameters are frozen throughout FPO training.

For the main anchored FPO setting, we use an output-distribution distance over the same auxiliary \rev{$(\epsilon_s,\tau_s)$} samples used by the FPO ratio. Let \rev{$f_{\theta}^{s}=f_\theta(x_{\tau_s},\tau_s,c)$} and \rev{$f_{\mathrm{fix}}^{s}$} denote the current and frozen-baseline \rev{clean-state} predictions for sample \rev{$s=1,\dots,N_{\mathrm{aux}}$}. The implementation supports three anchor forms:
\begin{align}
\mathcal{L}_{\mathrm{BC}}^{\mathrm{MSE}}
&= \frac{1}{|\Omega|\rev{N_{\mathrm{aux}}}}\sum_{\rev{s=1}}^{\rev{N_{\mathrm{aux}}}}\sum_{(u,d)\in\Omega}
\rev{\big(f_{\theta,u,d}^{s}-f_{\mathrm{fix},u,d}^{s}\big)^2},\\
\mathcal{L}_{\mathrm{BC}}^{\mathrm{W2}}
&= \frac{1}{|\Omega|}\sum_{(u,d)\in\Omega}
\left[
\rev{(\bar{f}_{\theta,u,d}-\bar{f}_{\mathrm{fix},u,d})^2}
+\rev{(\varsigma_{\theta,u,d}-\varsigma_{\mathrm{fix},u,d})^2}
\right],\\
\mathcal{L}_{\mathrm{BC}}^{\mathrm{KL}}
&= \frac{1}{2|\Omega|}\sum_{(u,d)\in\Omega}
\left[
\rev{\log\frac{\varsigma^2_{\mathrm{fix},u,d}}{\varsigma^2_{\theta,u,d}}}
+\rev{\frac{\varsigma^2_{\theta,u,d}+(\bar{f}_{\theta,u,d}-\bar{f}_{\mathrm{fix},u,d})^2}{\varsigma^2_{\mathrm{fix},u,d}}}
-1
\right],
\end{align}
where \rev{$\bar{f}$} and \rev{$\varsigma$} are \rev{the mean and standard deviation of $f^{s}$} across the auxiliary samples. The reported \textit{FPO w/o BC Anchor} ablation keeps the FPO surrogate-ratio machinery but removes this frozen-baseline output regularizer by setting the BC loss coefficient to zero. This isolates the effect of anchoring the planner to its supervised motion prior from the effect of the FPO update itself.

\subsubsection{Planner Critic Architecture}
\label{sec:appendix_fpo_critic}

\rev{The FPO baselines use a planner-level critic that estimates the state value before each reference window is executed. It is an MLP with LayerNorm and SiLU activations whose base input concatenates the planner's fixed prefix states, the current simulator state (root height, local body positions, rotations, and velocities, and contact indicators), and the task goals. Optional branches encode privileged object and goal features, the controller observation, the CLIP text embedding, the object BPS features, and the flow-matching conditions. The critic never observes planner samples or controller actions, and its value estimates serve as the GAE baseline of the surrogate-ratio update. SFPO does not use a critic, since group-relative normalization of the sequence scores replaces this learned baseline.}

\subsection{\texorpdfstring{\rev{Reward Terms and Group-Relative Advantage}}{Reward Terms and Group-Relative Advantage}}
\label{sec:appendix_group_advantage}

\rev{The reward terms of Sec.~\ref{sec:hoi_reward} take the forms}
\begin{equation}
\begin{gathered}
\rev{r^t_{\mathrm{humanoid}}=\exp\big(-\alpha_{\mathrm{h}}e^t_{\mathrm{h}}\big),\qquad r^t_{\mathrm{object}}=\exp\big(-\alpha_{\mathrm{o}}e^t_{\mathrm{o}}\big),}\\
\rev{R_{\mathrm{success}}=\mathbb{1}[\text{task success}],\qquad R_{\mathrm{term}}=-\mathbb{1}[\text{episode ends without success}],}
\end{gathered}
\end{equation}
\rev{where $e^t_{\mathrm{h}}$ is the mean per-joint position error of the humanoid and $e^t_{\mathrm{o}}$ is the object center position error at step $t$, and $\alpha_{\mathrm{h}},\alpha_{\mathrm{o}}>0$ are scale coefficients.}

Although the controller and the planner share the same reward components, they utilize these rewards differently. The controller operates as a stepwise policy. Consequently, its \rev{PPO} reward at each simulation step $t$ is formulated as
\begin{equation}
r_t^{\mathrm{ctrl}} = R_{\mathrm{trans}}^t + \mathbb{1}[t=T]R_{\mathrm{task}},
\end{equation}
where $T$ denotes the termination time of the current episode. The controller receives the tracking-based transition reward at every step and an additional task reward at the end of the episode.

\rev{Each planner sample is scored by the time-averaged sequence reward $R^{\mathrm{seq}}$ of Eq.~\eqref{eq:sequence_reward}, which reflects both the tracking feasibility of its references during closed-loop execution and whether the task is accomplished.}

In planner group-relative policy optimization, we evaluate sample quality via closed-loop controller and physics engine execution, rather than designing intermediate rewards for individual denoised reference windows \rev{$\hat{x}_{t:t+N_f-1}$}. An optimization group forms a semantically consistent reset cohort. Sharing identical subject-object pairs, reference indices, start times, contact reset states, and task conditions, group rollouts ensure uniform task semantics. Members differ solely through independent SFPO denoising noise, yielding multiple stochastic samples under identical interaction conditions. Given the $i$-th cohort condition $c_i$, we sample $G$ rollouts per group. The $g$-th rollout generates reference motion \rev{$\hat{x}^{\,i,g}_{t:t+N_f-1}$}. The controller executes the initial $N_r$ steps, continuing closed-loop replanning until success, failure, or timeout. \rev{The rollout then receives the sequence reward $R_i^g$, \emph{i.e.}, $R^{\mathrm{seq}}$ of Eq.~\eqref{eq:sequence_reward} evaluated on this rollout with its own length $T$.}

Excluding invalid or incomplete rollouts, we define the valid sample set as $\rev{\mathcal{V}_i} = \rev{\{g\in\{1,\dots,G\} \mid \text{rollout } g \text{ has terminated and } R_i^g \text{ is finite}\}}$. This yields the group mean and standard deviation:
\begin{equation}
\bar{R}_i = \frac{1}{|\rev{\mathcal{V}_i}|} \sum_{g \in \rev{\mathcal{V}_i}} R_i^g, \quad \rev{\varsigma_i} = \rev{\operatorname{std}} \left( \{R_i^g : g \in \rev{\mathcal{V}_i}\} \right),
\end{equation}
where \rev{$\varsigma_i$} is the unbiased sample standard deviation. Groups with $\rev{\varsigma_i}\le10^{-6}$ carry no relative preference and are skipped. For the remaining groups, we assign a clipped group-relative advantage to each valid rollout:
\begin{equation}
A_i^g = \operatorname{clip} \left( \frac{R_i^g - \bar{R}_i}{\max(\rev{\varsigma_i}, 10^{-12})}, -A_{\max}, A_{\max} \right).
\label{eq:clipped_advantage}
\end{equation}
\begin{algorithm}[t]
\caption{Coordinated planner-controller learning in InfCoiL.}
\label{alg:coordinated}
\begin{algorithmic}[1]
\Require planner $P_\theta$ pretrained by flow matching, controller $\pi_\phi$ from distillation and open-loop RL, minimum phase lengths $N_{\mathrm{c}}^{\min}$, $N_{\mathrm{p}}^{\min}$
\While{training budget not exhausted}
  \Statex \textit{Controller phase (planner frozen)}
  \Repeat
    \State Roll out in closed loop: replan with $P_\theta$ every $N_r$ steps and act with $\pi_\phi$
    \State Update $\pi_\phi$ with PPO on the HOI-centric reward (\rev{Sec.~\ref{sec:hoi_reward}})
  \Until{at least $N_{\mathrm{c}}^{\min}$ iterations and tracking quality or \rev{success rate} stops improving}
  \Statex \textit{Planner phase (controller frozen)}
  \Repeat
    \State $\theta_{\mathrm{old}}\gets\theta$
    \ForAll{task conditions $c_i$ in the batch}
      \For{$g=1,\dots,G$}
        \State Execute with $\pi_\phi$, replanning with $\pi_{\theta_{\mathrm{old}}}$ (Eq.~\eqref{eq:sfpo_policy}), until termination
        \State Store $\bar\ell_{\theta_{\mathrm{old}}}^{\,i,g,t}$ for every replanning event $t$ and compute $R_i^g$
      \EndFor
      \State Compute $A_i^g$ by Eq.~\eqref{eq:clipped_advantage}, skipping the group if its reward variance is near zero
    \EndFor
    \For{up to 5 epochs, with KL early stopping}
      \State Recompute $\bar\ell_{\theta}^{\,i,g,t}$ on the stored paths and update $\theta$ by ascending $\mathcal{J}_{\mathrm{SFPO}}$ (Eq.~\eqref{eq:sfpo_objective})
    \EndFor
  \Until{at least $N_{\mathrm{p}}^{\min}$ iterations and sequence score or \rev{success rate} stops improving}
\EndWhile
\end{algorithmic}
\end{algorithm}

\rev{Groups with fewer than $V_{\min}$ valid rollouts, which can occur when rollouts terminate at different times or batches are small, are also masked out. When a rollout contains several replanning events, its advantage $A_i^g$ is assigned to every valid replanning event, each forming one training sample, so that every generated reference window, not only the last one, receives credit for the episode outcome.}

\subsection{Details of Coordinated Learning Paradigm}
\label{sec:appendix_coordinated_learning}

\rev{InfCoiL adopts} a coordinated learning paradigm alternating between controller and planner adaptation. Before joint training, we pretrain the planner via supervised flow matching and derive a student controller via behavioral cloning distillation and open-loop reinforcement learning. Subsequently, we alternately optimize both components in the closed-loop environment. The system switches to planner reinforcement learning when controller tracking performance or success rate plateaus, and reverts to controller reinforcement learning when planner sequence-level success or rollout scores plateau. \rev{This alternation lets the controller adapt to the evolving planner distribution, while the planner can exploit the larger set of references that the improved controller can execute.}

During the controller adaptation phase, we fix the planner and collect closed-loop rollouts. The controller is updated with proximal policy optimization using the stepwise reward \rev{$r^{\mathrm{ctrl}}_t$}. \rev{We track moving averages of the tracking quality and the success rate together with their best values in the current phase. To prevent frequent switching, each controller phase lasts at least $N_{\mathrm{c}}^{\min}$ iterations (Algorithm~\ref{alg:coordinated}). After that, the system switches to the planner phase once either moving average fails to improve on its phase best by more than a tolerance $\delta_{\mathrm{c}}$.}

During the planner adaptation phase, the controller is fixed. The system collects grouped closed-loop rollouts by sampling references using SFPO noise. We compute group-relative advantages and update the planner accordingly. \rev{Analogously, we track moving averages of the sequence reward and the success rate, and after at least $N_{\mathrm{p}}^{\min}$ iterations the system returns to the controller phase once either fails to improve on its phase best by more than a tolerance $\delta_{\mathrm{p}}$.}

Algorithm~\ref{alg:coordinated} summarizes the procedure. In the reported experiments, the phase lengths are fixed rather than triggered by plateaus: InfCoiL (C$\to$P) uses 10k controller iterations followed by 500 planner iterations, and InfCoiL (C$\to$P$\to$C) adds 4k further controller iterations\rev{, so the tolerances $\delta_{\mathrm{c}}$ and $\delta_{\mathrm{p}}$ play no role in the reported results}.

\section{Additional Discussion}
\label{app:additional_discussion}

This section provides extended literature context, outlines future directions, and discusses broader impacts beyond the main paper.

\subsection{Extended Related Work}
\label{sec:appendix_related_works}
\textbf{Physics-based Humanoid Control.}
Physics-based humanoid control aims to produce physically plausible full-body motions by controlling simulated humanoids~\cite{peng2018deepmimic}. A common paradigm is to train RL policies to track motion capture datasets at scale, resulting in general-purpose motion trackers~\cite{luo2023perpetual, tessler2024maskedmimic}. Generalized behavior-cloning frameworks further retarget mocap data to diverse humanoid morphologies and learn whole-body imitation policies with DAgger-style distillation and PPO~\cite{yao2025gbc}.
Subsequent methods distill reusable motion priors from such trackers through adversarial imitation~\cite{peng2022ase,peng2021amp} or latent skill representations including VAEs and vector-quantized models~\cite{tessler2024maskedmimic,luo2023universal,yao2022controlvae,yao2024moconvq}. 
Recent diffusion-based controllers further model state-action trajectories for flexible task guidance~\cite{wu2025uniphys,huang2025diffuse}.
Furthermore, text- and task-conditioned control has also been explored through behavior cloning and multi-stage reinforcement learning~\cite{juravsky2024superpadl,tessler2023calm,truong2024pdp}, but these policies often provide limited motion diversity compared with kinematic generative models. 
To exploit the strengths of both paradigms, planner-controller frameworks combine diffusion-based kinematic planners with physics-based trackers~\cite{hansen2024hierarchical, rempe2023trace, tevet2024closd}.
However, these methods largely consider the humanoid alone, without explicitly modeling object dynamics or sustained contact-rich human-object coupling, which limits their applicability to physically grounded HOI tasks.

\textbf{Generative Planning with Closed-Loop Control and RL.}
Among planner-controller frameworks for humanoids, CLoSD~\cite{tevet2024closd} closes the loop by replanning from simulated states, but adapts only the tracking controller while the diffusion planner stays fixed. In robotics, AdaptDiffuser~\cite{liang2023adaptdiffuser} self-evolves a diffusion planner by \rev{generating synthetic trajectories, filtering them, and fine-tuning the planner on the retained ones}. Hierarchical diffusion-RL planning~\cite{wu2024diffusion} trains a low-level RL policy beneath a high-level diffusion path planner, and RLingua~\cite{chen2024rlingua} and AnyBipe~\cite{yao2025anybipe} use large language models to guide RL through generated controllers or reward design rather than training a generative motion planner. A separate line fine-tunes diffusion and flow policies with online RL, via stochastic denoising chains~\cite{ren2025diffusion, zhang2025reinflow, liu2025flow} or flow-matching surrogate likelihoods~\cite{mcallister2025flow, yi2026flow}. In these settings, generated samples are scored directly by the environment or a reward model. InfCoiL differs in which component is optimized by execution feedback: it updates the flow-matching planner on-policy from complete closed-loop rollouts, where each plan is scored only after being executed by a separately learned controller under contact-rich dynamics, and alternates this with controller adaptation so that both sides of the planner-controller interface move together.

\textbf{Generative Models for Human-Object Interaction.}
Large-scale 4D human-object interaction (HOI) datasets provide paired human-object motion and geometry annotations for learning interaction generators~\cite{taheri2020grab, bhatnagar2022behave, li2023object, lu2025humoto}. A common direction is to enrich diffusion models with object motion, contact, language, geometry, and waypoint conditions, enabling physics-informed prediction, object-guided manipulation, and text-controllable interaction synthesis~\cite{xu2023interdiff, li2023object, diller2024cg, li2024controllable}. Later methods further strengthen the conditioning structure with semantic-geometric context, object relations, kinematic-chain representations, unseen-object generalization, and interaction dynamics~\cite{cong2025semgeomo, xue2025guiding, zeng2025chainhoi, li2025genhoi, wu2025hoi}. Recent studies also revisit the guidance and denoising mechanism itself: \cite{wang2026unleashing} obtains interaction guidance without auxiliary classifiers by asynchronously denoising body, hand, and object components, and introduces human shape as conditions. Some methods have explored consistency models for motion generation to accelerate inference, which is crucial for online planning~\cite{dai2024motionlcm, jiang2025motionpcm, zou2026infbagel}. However, flow-based HOI generation remains underexplored. Moreover, existing generators are typically trained and sampled without physical execution feedback, limiting their ability to adapt plans to the states actually reached by the controller.

\textbf{Physics-based Human-Object Interaction.}
Physics-based HOI methods aim to execute interactions under contact-rich dynamics rather than only synthesize kinematic trajectories. A direct paradigm is to learn task policies with reinforcement learning from manually designed rewards, which can produce executable behaviors for sports, games, and object manipulation but often depends on task-specific engineering and lacks demonstration-level realism~\cite{merel2020catch, liu2018learning, xu2024humanvla, gao2024coohoi, pan2025tokenhsi}. Imitation-based methods reduce this burden by tracking human-object demonstrations in simulation, enabling contact-rich and reusable interaction skills~\cite{wang2023physhoi, wang2025skillmimic, yu2025skillmimic}. However, these skills remain bounded by the objects, motions, and interaction modes covered by demonstrations, limiting their ability to synthesize diverse long-horizon behaviors from fine-grained textual and spatial instructions. Recent large-scale methods improve coverage by learning whole-body controllers from broader and imperfect mocap interaction data, including unified controllers for heterogeneous objects and sparse-intention-conditioned control~\cite{xu2025intermimic, xu2026interprior}. These methods ground HOI in simulation, but they remain less expressive than text-conditioned generative planners in semantic controllability and motion diversity. Hierarchical systems further combine interaction generators with imitation policies to connect planning and physical tracking~\cite{lin2025simgenhoi}, but they still generate static references before tracking, leaving the planner essentially open-loop. Our work addresses this gap with a closed-loop planner-controller framework, where the planner repeatedly conditions on language, goals, morphology, object states, and recent physical feedback, and coordinated reinforcement learning adapts both modules to each other.

\subsection{Future Directions}
\label{app:future_directions}
Beyond the limitations discussed in Sec.~\ref{sec:conclusion}, several directions could extend InfCoiL. First, the planner currently consumes structured object states and goal specifications. Replacing them with perception-driven inputs would connect InfCoiL to vision-driven embodied agents that explore and act in open-world scenes~\cite{mi2026exploratory}, and to physical world models and world action models that jointly predict visual dynamics and actions~\rev{\cite{cai2026internw0, miao2026internw0, zou2026picawamvisualperturbationinvariance}}, which could also provide learned dynamics priors beyond the kinematic mocap prior used by our planner. Second, the closed-loop planner-controller coordination could be evaluated beyond whole-body HOI, for example on bimanual and mobile robotic manipulation benchmarks with diverse objects and tasks~\cite{mu2024robotwin, mu2025robotwin, chen2025robotwin, gao2026ebench}\rev{, where coordinated learning would carry over through domain-specific execution rewards and planner-controller interface definitions}. Third, transferring the multi-morphology controller to physical humanoids would require \rev{accurate real-time tracking of the object states consumed by the planner,} cross-embodiment motion retargeting and imitation~\cite{yao2025gbc} as well as automated sim-to-real training and evaluation pipelines~\cite{yao2025anybipe}, together with adaptation to hardware constraints, sensing noise, and real contact dynamics. \rev{Robot platforms such as Unitree G1 lie far outside the SMPL-X shape variation used in our experiments, so preserving the same interaction diversity would also require dynamics-aware adjustment of proportions, joints, contacts, balance, and actuation. Fourth, InfCoiL could serve as an executable skill for language-model agents that write control programs~\cite{liang2023code}, invoke learned skills~\cite{ahn2022can}, or evolve reward programs for humanoid loco-manipulation~\cite{lin2026interevolve}. A recent EBench study finds that such agents benefit from invoking post-trained policies but are limited by fixed-instruction execution without subgoals~\cite{yao2026profiling}, a gap that the instruction- and goal-conditioned replanning of InfCoiL could fill. Fifth, the injected semantic noise is small, with final-step standard deviations below $10^{-2}$ in normalized units, so numerical discrepancies between rollout and update passes, e.g., from reduced-precision arithmetic~\cite{micikevicius2018mixed, kalamkar2019study}, can be amplified in the log-probability ratio. Since low precision has been linked to loss spikes~\cite{liu2026grokking} and to training-inference mismatch in RL fine-tuning~\cite{qi2025defeating}, the numerical precision and the hand-set noise scales $\kappa$ deserve a joint study. Finally, the stochastic-flow construction of SFPO does not depend on the motion representation and may extend to other flow-matching diffusion transformer (DiT) planners, in particular Mixture-of-Transformers variants with heterogeneous modality streams, for which the semantic covariance can be allocated per stream.}

\subsection{Broader Impact}
InfCoiL studies text- and goal-conditioned physics-based human-object interaction for controllable character animation and \rev{simulated embodied agents}. The proposed framework may benefit applications such as animation authoring, virtual character control, robotics simulation, assistive embodied agents, and synthetic data generation, by reducing the manual effort required to design physically plausible interaction behaviors. At the same time, improving the controllability and realism of simulated human-object interactions may raise potential risks if such techniques are used to generate misleading visual content, imitate human behaviors without proper disclosure, or transfer unsafe interaction policies to real robotic systems. Our work is conducted in simulation and focuses on physically grounded interaction modeling rather than human identity replication. We encourage responsible use of the released code and models, including clear disclosure when generated motions are used in media applications, careful safety validation before any real-world robotic deployment, and respect for the licenses and consent conditions of the motion and object datasets used for training and evaluation.

\end{document}